%% file: iclr2027_conference.tex
\documentclass{article} 
\usepackage{iclr2027_conference,times}

\input{math_commands.tex}

\usepackage{hyperref}
\usepackage{url}
\usepackage{microtype}
\usepackage{booktabs} 
\usepackage{amsmath}
\usepackage{amssymb}
\usepackage{mathtools}
\usepackage{amsthm}
\usepackage{algorithm}
\usepackage{algorithmic}

\usepackage{enumitem}
\usepackage{threeparttable}
\usepackage{multirow}
\usepackage{rotating}
\usepackage{makecell}
\usepackage{amsfonts}
\usepackage{flushend}
\usepackage{color}
\usepackage{bm}
\usepackage{subcaption}
\usepackage{bbm}
\usepackage{pifont}
\usepackage[most]{tcolorbox}

\usepackage[capitalize,noabbrev]{cleveref}
\usepackage{wrapfig}
\usepackage{booktabs}
\usepackage{graphicx}
\usepackage{xcolor}
\usepackage{multirow}
\usepackage{colortbl}
\usepackage[normalem]{ulem}
\usepackage{xspace}

\definecolor{win1}{rgb}{ .439,  .678,  .278}
\definecolor{win2}{rgb}{ .267,  .447,  .769}
\definecolor{win3}{rgb}{ .929,  .49,  .192}
\definecolor{graybg}{gray}{0.9}      

\newcommand{\oursyn}{\textsc{AG-Forge}\xspace}
\newcommand{\ourmethod}{\textsc{TS-GGAD}\xspace}

\newcommand{\first}[1]{\textcolor{win1}{\textbf{#1}}}
\newcommand{\second}[1]{\textcolor{win2}{#1}}
\newcommand{\third}[1]{\textcolor{win3}{#1}}

\definecolor{challengeColor}{HTML}{8E3B46}
\definecolor{contributionColor}{HTML}{2F5D50}
\definecolor{agforgePurple}{rgb}{0.93,0.90,0.97}

\title{Is Real-World Training Data Necessary for Generalist Graph Anomaly Detection?}

\author{
Yujing Liu$^{1}$\thanks{Equal contribution.},
Yixin Liu$^{1}$\footnotemark[1],
Yue Tan$^{1}$,
Xiaofeng Cao$^{2}$,
Alan Wee-Chung Liew$^{1}$,
Heng Tao Shen$^{2}$,
Shirui Pan$^{1}$\thanks{Corresponding author.} \\
$^{1}$School of Information and Communication Technology, Griffith University, Australia \\
$^{2}$School of Computer Science and Technology, Tongji University, China \\
\texttt{yujing.liu@griffithuni.edu.au} \quad
\texttt{yixin.liu@griffith.edu.au} \\ \texttt{s.pan@griffith.edu.au}
}

\iclrfinalcopy 
\begin{document}

\maketitle
\fancyhead[L]{Preprint}

\begin{abstract}

Generalist graph anomaly detection (GAD) aims to build a foundation model that 
detects anomalies on arbitrary unseen graphs without retraining or fine-tuning. 
Sufficient data are essential for foundation model training, yet generalist GAD still faces a data shortage, as real-world anomalous graphs are scarce and costly to collect and annotate. 
To fill this gap, we propose \textbf{\oursyn}, an \textbf{\underline{A}}nomalous \textbf{\underline{G}}raph generation \textbf{\underline{Forge}} for automatic synthesis of anomalous graphs, exploring the feasibility of synthetic data-driven training for generalist GAD. Empirically, we find that synthetic data can achieve performance comparable to real-world training, but fail to push the performance boundary further due to the limited capacity of existing methods. 
To further unlock model capacity as training data scale up, we develop \textbf{\ourmethod}, a \textbf{\underline{T}}opology-\textbf{\underline{S}}emantic coordinated \textbf{\underline{G}}eneralist \textbf{\underline{GAD}} that captures complementary topological and semantic anomaly evidence, together with a curriculum learning strategy tailored to large-scale synthetic training. 
Extensive experiments on 14 real-world datasets demonstrate that \ourmethod, trained on data generated by \oursyn, significantly outperforms state-of-the-art methods. 

\end{abstract}

\section{Introduction}
\input{Section/intro}

\section{Preliminaries and Related Work}
\input{Section/RW}

\section{Methodology}
\input{Section/method}

\section{Experiments}
\input{Section/experiment}
\section{Conclusion}
\input{Section/conclusion}

\bibliography{iclr2027_conference}
\bibliographystyle{iclr2027_conference}

\appendix

\clearpage
\input{Section/appendix}

\end{document}

%% file: math_commands.tex
\usepackage{amsmath,amsfonts,bm}

\def\eqref#1{equation~\ref{#1}}

\def\1{\bm{1}}

\DeclareMathAlphabet{\mathsfit}{\encodingdefault}{\sfdefault}{m}{sl}
\SetMathAlphabet{\mathsfit}{bold}{\encodingdefault}{\sfdefault}{bx}{n}



%% file: Section/intro.tex
Graph anomaly detection (GAD) aims to identify abnormal entities in graph-structured data, with applications in a wide range of real-world scenarios, such as financial fraud detection, network intrusion detection, and social security monitoring~\cite{qiao2025deep}. Conventional GAD methods require dataset-specific training on each target graph, leading to high deployment costs and poor generalizability in new scenarios~\cite{ding2019deep,liu2021anomaly}.  To address these limitations, recent studies began to explore \textbf{generalist GAD} approaches that learn transferable anomaly patterns from diverse graphs, enabling a single GAD model to generalize to unseen graphs without dataset-specific retraining or fine-tuning~\cite{liu2024arc,zhang2026ia}. By reducing deployment costs and extending GAD to broader application scenarios, generalist GAD represents an important step toward anomaly detection foundation models for graphs~\cite{qiao2025anomalygfm}.

For foundation models, sufficient and high-quality training data are essential to support cross-domain generalization and transferability~\cite{kaplan2020scaling,hollmann2025accurate}. Generalist GAD is no exception: ample and diverse training data from different scenarios are critically needed to capture a broad spectrum of anomaly patterns~\cite{liu2026rethinking}. 
However, unlike language and image data that are widely available at scale, real-world training data for GAD are much harder to collect -- Anomalies are inherently rare, while collecting large-scale graph data can be costly and subject to privacy or access constraints. As a result, current generalist GAD models are merely trained on 1-7 datasets from limited domains, with a total of 90k nodes in average~\cite{niu2024zero,qiao2025anomalygfm}. Such \textbf{data drought}, consequently, may become a key bottleneck to scaling up generalist GAD and extending its generalization to broader domains. %

A promising solution for data drought is to generate training data through \textbf{data synthesis}. Recent breakthroughs in PFN-style models demonstrate that foundation models can be trained entirely on synthetic tabular and time series data~\cite{hollmann2025accurate,dooley2023forecastpfn}. Using data synthesis, training data can be generated at virtually unlimited scale, alleviating the scarcity of real-world data; more importantly, their distributions and patterns are controllable, reducing the risk of low-quality data with noise and bias. 
Given the potential of synthetic data and the shortage of real-world data, a natural question arises: \textit{\underline{Is real-world training data necessary for generalist GAD?}}

\begin{wrapfigure}{r}{0.35\textwidth}
  \includegraphics[width=0.35\textwidth]{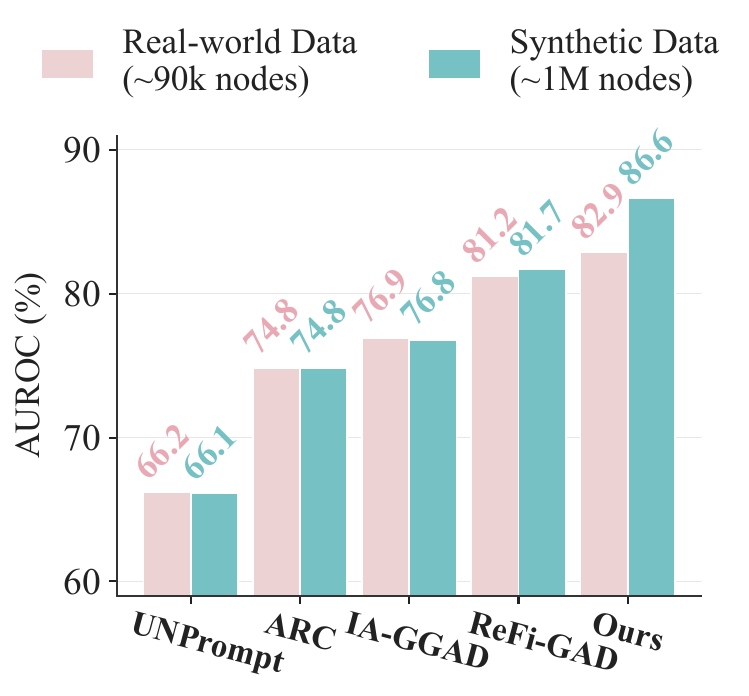}
    \vspace{-8mm}
  \caption{Real-world vs. synthetic data for model training.}
  \label{fig.real_syn}
    \vspace{-4mm}
\end{wrapfigure}

\textcolor{challengeColor}{\textbf{Challenge~\ding{182}:~Fidelity of synthetic graphs and anomalies.}} 
While data synthesis offers a promising pathway, generating training data for GAD remains non-trivial due to the inherent complexity of real-world graphs and anomaly patterns. \textit{On the one hand}, real-world graphs exhibit strong data heterogeneity, with graphs from different domains often showing distinct structure and feature distributions. \textit{On the other hand}, graph anomalies show highly complex and diverse patterns, with different real-world scenarios giving rise to distinct anomalous behaviors and interaction patterns. 
The coupling among graph structure, node features, and anomaly patterns makes it difficult to comprehensively simulate real-world anomalous graphs using simple generative models or heuristic rules. 

\textcolor{contributionColor}{\textbf{Contribution~\ding{182}:~\oursyn for anomalous graph synthesis.}} 
To address the above challenge, we design \textbf{\oursyn}, an \textbf{\underline{A}}nomalous \textbf{\underline{G}}raph generation \textbf{\underline{Forge}} for automatic synthesis of anomalous graphs. With generality and flexibility as key design goals, \oursyn employs a three-stage cascaded pipeline to synthesize anomalous graphs. Specifically, we incorporate four structural generation strategies and three feature synthesis schemes to construct graphs with diverse topological and semantic characteristics, together with anomaly injection mechanisms that simulate different anomalous behaviors. 
These components can be flexibly composed and varied, allowing \oursyn to generate a virtually unlimited variety of anomalous graphs.
Moreover, \oursyn manipulates anomaly difficulty through perturbation strength and camouflage, producing anomalies with controllable levels of detectability. Trained solely on the synthetic data generated by \oursyn, existing generalist GAD methods achieve comparable or even slightly better performance than those trained on real-world data (see Fig.~\ref{fig.real_syn}), demonstrating the viability of fully synthetic training for generalist GAD.

\textcolor{challengeColor}{\textbf{Challenge~\ding{183}:~Model and training bottlenecks at scale.}} 
With data generated by \oursyn for training, existing generalist GAD methods show no significant performance gains (Fig.~\ref{fig.real_syn}) from scaling the training data to about 1M nodes. This exposes a new bottleneck: as training data scale up, model capacity becomes the limiting factor. Due to the limited amount and diversity of available training data, existing generalist GAD approaches are typically built with limited model complexity and expressiveness, restricting their ability to capture diverse anomaly patterns in complex synthetic graph environments. Moreover, overwhelming synthetic data with varying detection difficulty may challenge conventional uniform training strategies, potentially leading to suboptimal learning. 

\textcolor{contributionColor}{\textbf{Contribution~\ding{183}:~Scalable generalist GAD model with curriculum learning.}} 
To unlock the potential of large-scale synthetic training, we develop \textbf{\ourmethod}, a more powerful generalist GAD model. \ourmethod explicitly captures anomaly-relevant \textbf{\underline{T}}opological patterns and \textbf{\underline{S}}emantic cues for \textbf{\underline{G}}eneralist \textbf{\underline{GAD}} through two dedicated streams, with an adaptive module coordinating their contributions for anomaly detection. 
Furthermore, to effectively exploit the large-scale synthetic corpus, we design a \textbf{curriculum learning strategy} that progressively introduces graphs of increasing difficulty while placing greater emphasis on challenging samples. Trained purely on large-scale synthetic graphs and evaluated on 14 real-world datasets, \ourmethod achieves the best performance on all datasets, with an average AUROC improvement of 5.39\% over the best baseline.

%% file: Section/RW.tex
\noindent\textbf{Notations and Problem Formulation.} Let an (anomalous) graph be $\mathcal{G}=(\mathcal{V},\mathcal{E},\mathbf{A},\mathbf{X})$, where $\mathcal{V}$ denotes the node set, $\mathcal{E}$ denotes the edge set represented by an adjacency matrix $\mathbf{A}\in\{0,1\}^{N\times N}$, and $\mathbf{X}\in\mathbb{R}^{N\times D}$ is the node feature matrix, with $N=|\mathcal{V}|$. In the context of graph anomaly detection (GAD), $\mathbf{y}\in\{0,1\}^{N}$ denotes the anomaly labels of $\mathcal{G}$, where ${y}_i=1$ indicates that the $i$-th node is anomalous, while ${y}_i=0$ indicates that it is normal. A GAD model $f(\cdot)$ aims to predict node-wise anomaly scores $\mathbf{\hat{y}}$, i.e., $f(\mathcal{G})=\mathbf{\hat{y}}$, where ${\hat{y}}_i$ indicates the abnormality of the $i$-th node: a larger ${\hat{y}}_i$ implies a higher likelihood of being anomalous.

\noindent\textbf{Dataset-Specific GAD Methods.} 
Conventional GAD methods typically follow a dataset-specific learning paradigm~\cite{qiao2025deep}. Formally, the dataset-specific GAD model $f(\cdot)$ learns anomaly patterns from a specific graph $\mathcal{G}$ through either unsupervised objectives, such as reconstruction~\cite{ding2019deep} and contrastive learning~\cite{liu2021anomaly}, or supervised binary classification that distinguishes normal and anomalous nodes with labeled training data~\cite{tang2022rethinking}. The learned GAD model $f(\cdot)$ is then applied to the same target graph $\mathcal{G}$ for anomaly detection. A major {limitation} of this dataset-specific learning paradigm is that it requires retraining for each new graph, resulting in high deployment costs and limited generalizability to unseen graphs and domains.

\noindent\textbf{Generalist GAD Methods: The Foundation Models for GAD.} 
Inspired by the success of foundation models, emerging studies have explored generalist GAD methods that learn transferable anomaly knowledge from multiple source graphs and generalize to unseen graphs without graph-specific retraining or fine-tuning~\cite{liu2024arc}. Formally, the generalist GAD model $f(\cdot)$ is trained on a collection of source graphs $\mathcal{D}_{sr}=\{\mathcal{G}_{1}^{s}, \cdots,\mathcal{G}_{n}^{s}\}$ to learn transferable anomaly patterns; after training, the generalist model $f(\cdot)$ can be directly applied to predict anomaly scores on any unseen target graph $\mathcal{G}^t\in\mathcal{D}_{tg}$ from arbitrary target domains, where $\mathcal{D}_{sr} \cap \mathcal{D}_{tg}=\emptyset$. 

To achieve domain-agnostic generalization, existing methods place strong emphasis on the design of generalist GAD models for transferable anomaly detection~\cite{qiao2025anomalygfm,zhang2026ia}. For instance, ARC~\cite{liu2024arc} uses in-context learning for test-time adaptation, UNPrompt~\cite{niu2024zero} adopts learnable prompts for cross-graph transfer, and ReFi-GAD~\cite{liu2026rethinking} learns unified embeddings for cross-domain generalization. Nevertheless, the limited real-world data available for GAD model training, with only about 90k nodes forming $\mathcal{D}_{sr}$, may pose a potential bottleneck to scaling up generalist GAD and improving its generalization capability.

\noindent\textbf{Data Synthesis for Foundation Models.}
To alleviate data shortage, a growing line of research explores synthetic data generation for foundation model training. A representative example is TabPFN, which pretrains a transformer-based foundation model on million-level purely synthetic tabular tasks, enabling training-free in-context prediction on unseen tabular datasets~\cite{hollmann2025accurate}. Building on the success of PFNs, synthetic data have been used to train foundation models not only for tabular data~\cite{qu2025tabicl} but also for time series~\cite{taga2025timepfn} and relational data~\cite{wang2026relational}. Inspired by this advance, we explore data synthesis for generalist GAD by constructing $\mathcal{D}_{sr}= \mathcal{D}_{syn} = \{\mathcal{G}_i^s\}_{i=1}^{n}$, where each synthetic graph $\mathcal{G}_i^s \sim p_{\mathrm{syn}}(\mathcal{G})$ is sampled from a designed graph synthesis distribution.

A more comprehensive review of the related literature is provided in Appendix~\ref{appe:related_work}.

%% file: Section/method.tex
\begin{figure*}[t!]
    \centering
    \includegraphics[width=\textwidth,]{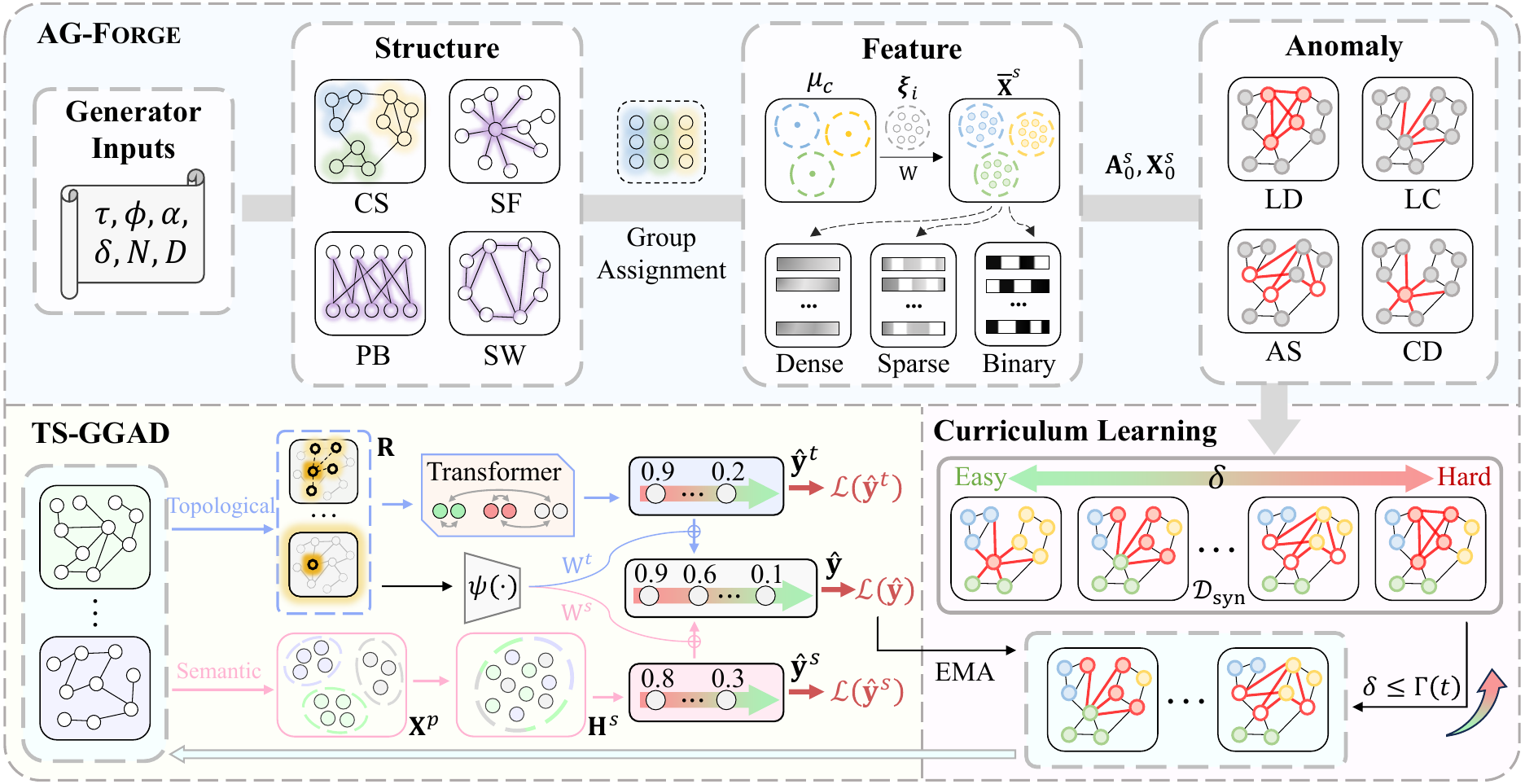}
    \caption{The framework of the proposed \textbf{\oursyn}, \textbf{\ourmethod}, and curriculum learning strategy.}\label{fig:framework} 
\end{figure*}

In this section, we first introduce \textbf{\oursyn}~(Section~\ref{Synthetic}), an anomalous data synthesis framework for generalist GAD, which constructs training graphs with diverse structures, feature patterns, and anomaly patterns. To fully leverage the rich synthetic data, we further develop \textbf{\ourmethod}~(Section~\ref{model}), a generalist anomaly detector to capture complementary topological and semantic anomaly evidence. Finally, we introduce a \textbf{curriculum learning} strategy~(Section \ref{Curriculum}) to guide the model in learning transferable anomaly knowledge from rich and diverse synthetic data. The overall framework of \oursyn, \ourmethod, and the proposed curriculum learning strategy is shown in Fig.~\ref{fig:framework}.

\subsection{\oursyn for Anomalous Graph Synthesis}\label{Synthetic}

To alleviate the data drought faced by generalist GAD, we design \textbf{\oursyn}, an \textbf{A}nomalous \textbf{G}raph generation \textbf{Forge} for automatic synthesis of anomalous graphs, aiming to generate scalable and diverse training data for generalist GAD. Given the strong heterogeneity of real-world graphs, our goal is to construct a synthetic space spanning diverse graph environments and anomaly patterns covering varying difficulty levels. To this end, we decouple the generation process into a three-stage cascaded pipeline: \textcolor{challengeColor}{\textbf{graph structure generation} \ding{220} \textbf{node feature generation} \ding{220} \textbf{anomaly injection}}.
Let $\tau$, $\phi$, and $\alpha$ denote the structure, feature, and anomaly types, respectively, while $\delta$ controls the anomaly difficulty. The overall generation process is formulated as
\begin{equation}
\begin{cases}
\mathbf{A}_{0}^{s}
= g_{\mathrm{str}}(N;\tau),
& \text{Step 1: Structure Generation}, \\

\mathbf{X}_{0}^{s}
= g_{\mathrm{feat}}(\mathbf{A}_{0}^{s},D;\phi),
& \text{Step 2: Feature Generation}, \\

(\mathbf{A}^{s},\mathbf{X}^{s},\mathbf{Y}^{s})
= g_{\mathrm{ano}}(\mathbf{A}_{0}^{s},\mathbf{X}_{0}^{s};\alpha,\delta),
& \text{Step 3: Anomaly Injection},
\end{cases}
\label{eq:syn_generation}
\end{equation}
where $N$ and $D$ denote the number of nodes and feature dimensionality, respectively, $\mathbf{A}_{0}^{s}$ and $\mathbf{X}_{0}^{s}$ represent the graph topology and node attributes before anomaly injection, while $\mathbf{A}^{s}$, $\mathbf{X}^{s}$, and $\mathbf{Y}^{s}$ denote the final anomalous topology, node attributes, and anomaly labels. By varying these generation factors, we construct diverse graphs varying in scale, structure, feature patterns, and anomaly patterns. More detailed formulations of each step in \oursyn are provided in Appendix~\ref{app:oursyn}.

\noindent\textbf{Step 1:~Structure Generation.}
To simulate the structural diversity of real-world graphs, we design four topology generation mechanisms with distinct structural characteristics. Given the randomly sampled graph size $N$ and structure type $\tau$, the base graph structure $\mathbf{A}_{0}^{s}$ is generated as
\begin{equation}
N \sim \mathcal{U}_{\mathbb{Z}}(N_{\min},N_{\max}), \qquad
\tau \sim \mathcal{U}\{\mathrm{CS},\mathrm{SF},\mathrm{PB},\mathrm{SW}\}, \qquad
\mathbf{A}_{0}^{s}=g_{{str}}(N;\tau).
\end{equation}
where $N$ is uniformly sampled from $[N_{\min}, N_{\max}]$, $\tau$ denotes the sampled structure type, and $\mathrm{CS}$, $\mathrm{SF}$, $\mathrm{BP}$, and $\mathrm{SW}$ denote community-structured, scale-free, projected bipartite, and small-world graphs, respectively. The characteristics of these structure models are summarized as follows:

\begin{itemize}[leftmargin=10pt, itemsep=0pt, parsep=0pt, topsep=0pt]
    \item \textbf{Community-structured graph (CS)} follows the stochastic block model (SBM)~\cite{abbe2018community}, forming explicit communities where intra-community connections are denser than inter-community ones. Varying community sizes and connection probabilities yields diverse modular structures.

    \item \textbf{Scale-free graph (SF)} captures hub-dominated connectivity patterns, where a small number of nodes have substantially higher degrees than others. Different growth and clustering settings produce diverse degree distributions and connectivity patterns.

    \item \textbf{Projected bipartite graph (PB)} models homogeneous relations from heterogeneous interactions. We construct a sparse bipartite graph between target and auxiliary nodes and project it onto the target node set, producing connectivity patterns determined by shared auxiliary neighbors.

    \item \textbf{Small-world graph (SW)} constructs a locally connected ring lattice and randomly rewires a fraction of edges to introduce long-range connections, producing high local clustering and short global paths. Such patterns are common in social and communication networks.
\end{itemize}

These four mechanisms produce the base topology $\mathbf{A}_{0}^{s}$ with diverse structural characteristics, providing a structural basis for subsequent anomaly injection. 
To guide subsequent feature generation, each node is assigned a group label $c_i$ denoting the community it belongs to. For CS graphs, $c_i$ follows the {SBM block assignment}; for other types of graphs, groups are obtained either by {Louvain community detection}~\cite{de2011louvain} or random partitioning, simulating different community patterns.

\noindent\textbf{Step 2:~Feature Generation.}
Apart from topology, real-world graphs also exhibit heterogeneity in node attributes. Accordingly, we construct three types of node attributes: dense continuous, sparse high-dimensional, and binary features. Given the base topology $\mathbf{A}_{0}^{s}$, we first construct base features according to the node groups $c$ and then transform them into different feature types. Formally,
\begin{equation}
D \sim \mathcal{U}_{\mathbb{Z}}(D_{\min},D_{\max}), \qquad
\phi \sim \mathcal{U}\{\mathrm{Dense},\mathrm{Sparse},\mathrm{Binary}\}, \qquad
\mathbf{X}_{0}^{s}=g_{\mathrm{feat}}(\mathbf{A}_{0}^{s},D;\phi).
\end{equation}
\textbf{\textit{Step 2.1:~Base feature generation.}}
We first assign each group $c$ a low-dimensional latent center $\boldsymbol{\mu}_c\in\mathbb{R}^{d_z}$. For each node $v_i$, we combine its group center with a node-specific variation $\boldsymbol{\xi}_i$ sampled from a shared mixture distribution, i.e., $\mathbf{z}_i = \boldsymbol{\mu}_{c_i} + \boldsymbol{\xi}_i$, and then project the result to the target feature dimension:~$\overline{\mathbf{x}}^s_{i} = \mathbf{z}_i\mathbf{W} + \boldsymbol{\epsilon}_i$,
where $\mathbf{W}\in\mathbb{R}^{d_z\times D}$ is a random projection matrix and $\boldsymbol{\epsilon}_i$ is gaussian noise. This design preserves group-level similarity and node-specific variation while introducing dependencies across feature dimensions, reflecting real-world feature distributions.

\textbf{\textit{Step 2.2:~Feature-type transformation.}}
After generating base features, we further transform them into one of three feature types: \textbf{Dense features} retain the original representation ($\mathbf{x}_{i}^{s} = \overline{\mathbf{x}}^s_{i}$), \textbf{sparse features} apply random sparsification ($\mathbf{x}_{i}^{s} = \mathrm{ReLU}(\overline{\mathbf{x}}^s_{i})\odot\mathbf{m}_i$), and \textbf{binary features} discretize each dimension into a binary value ($\mathbf{x}_{i}^{s} = \mathbb{I}\left(
\sigma(\overline{\mathbf{x}}_{i}^{s})>\boldsymbol{\tau}_i
\right), \ 
\boldsymbol{\tau}_i\sim U(0,1)$).
Here, $\mathbf{m}_i$ is a random binary mask controlling sparsity, $\sigma(\cdot)$ denotes the sigmoid function, and $\boldsymbol{\tau}_i$ is sampled from $U(0,1)$. By varying the latent distribution, projection, and feature type, we generate diverse feature patterns with different sparsity and statistics, enriching attribute environments for generalist GAD training.

\noindent\textbf{Anomaly Injection.}
After constructing the base structure and node features, we further inject anomalies with different patterns to simulate diverse abnormal behaviors in different graph scenarios. Given the anomaly type $\alpha$ and difficulty factor $\delta$, the injection process is formulated as
\begin{equation}
\alpha \sim \mathcal{U}\{\mathrm{LD},\mathrm{LC},\mathrm{AS},\mathrm{CD},\mathrm{Mix}\}, \quad
\delta \sim \mathcal{U}(\delta_{\min},\delta_{\max}), \quad
(\mathbf{A}^{s},\mathbf{X}^{s},\mathbf{Y}^{s})
= g_{\mathrm{ano}}(\mathbf{A}_{0}^{s},\mathbf{X}_{0}^{s};\alpha,\delta),
\end{equation}
where $\mathrm{LD}$, $\mathrm{LC}$, $\mathrm{AS}$, $\mathrm{CD}$ and $\mathrm{Mix}$ denote different anomaly injection strategies, specified as:

\begin{itemize}[leftmargin=10pt, itemsep=0pt, parsep=0pt, topsep=0pt]
    \item \textbf{Local densification (LD)} forms anomalous groups with unusually dense internal connections and shared feature patterns. As $\delta$ increases, anomalous nodes become more densely interconnected and form more coherent groups, making them increasingly similar to normal communities.

    \item \textbf{Link concentration (LC)} connects anomalous nodes to a small set of common targets while assigning them attributes inconsistent with these targets. Increasing $\delta$ reduces the feature deviation, making the anomalies less distinguishable.

    \item \textbf{Attribute suppression (AS)} connects anomalous nodes to randomly sampled targets while suppressing their attributes, e.g., by setting many feature values to zero. With larger $\delta$, more feature variation is introduced, making the suppression less apparent.

    \item \textbf{Community dispersion (CD)} rewires anomalous nodes to different parts of the graph and perturbs their attributes around the global feature pattern. As $\delta$ increases, the feature deviation decreases, making these anomalies more difficult to distinguish.

    \item \textbf{Mixed anomaly (Mix)} partitions the anomalous node set into several subsets and applies different injection strategies to them, allowing multiple anomaly patterns to coexist within the same graph.
\end{itemize}

By organizing the three synthesis stages into a unified ``forge'', \oursyn flexibly configures and combines graph structures, features, and anomaly patterns, covering a broad range of anomalous graph scenarios. Their flexible combinations can produce a virtually unlimited variety of anomalous graphs, providing rich training data for generalist GAD models.

\subsection{Topology-Semantic Coordinated Generalist GAD Model}\label{model}

As the data bottleneck is alleviated, we observe that model expressiveness emerges as a new bottleneck for generalist GAD (Fig.~\ref{fig.real_syn}). 
Existing generalist GAD methods are typically designed for small-scale training settings with only a limited number of real-world graphs. Hence, they usually adopt relatively simple architectures, which may struggle to capture diverse anomaly patterns from massive synthetic training data. To unlock the potential of the large-scale synthetic corpus, we develop \textbf{\ourmethod}, a more powerful \textbf{T}opology-\textbf{S}emantic coordinated \textbf{G}eneralist \textbf{GAD} model with three components: topological pattern encoding, semantic pattern encoding, and anomaly evidence coordination.

\noindent\textbf{Topological Pattern Encoding.}
Anomalous nodes often exhibit irregular neighborhood relations or atypical structural patterns, such as low neighborhood consistency. Inspired by~\cite{liu2026rethinking}, we characterize each node using five anomaly-aware topological statistics: \textit{global cosine similarity}, \textit{neighborhood cosine similarity}, \textit{neighborhood Euclidean distance}, \textit{degree}, and \textit{local clustering coefficient} (Details are in Appendix~\ref{app:topo}). For node $v_i$, its topological descriptor is defined as
\begin{equation}
\mathbf{r}_i=f_{\mathrm{top}}(v_i,\mathbf{A})\in\mathbb{R}^{5},
\label{eq:topology}
\end{equation}
where $f_{\mathrm{top}}(\cdot)$ denotes the topological description function. Each dimension is rank-normalized to reduce cross-graph scale variations and retain anomaly-indicative structural information.

Since the anomaly indication of the same topological pattern may vary across graphs, we use a small target-domain support set to contextualize the representations, with an attention mask distinguishing support and query nodes. Specifically, the Transformer-based encoder can be written as:
\begin{equation}
\mathbf{H}^{t}
=\operatorname{Transformer}([\mathbf{r}_1,\cdots,\mathbf{r}_N];\mathbf{M}),
\label{eq:top_context}
\end{equation}
where $\mathbf{M}$ is an attention mask that restricts each query node (i.e., the nodes to be predicted) to attend only to the support nodes (i.e., the nodes with given labels at test time) and itself. Finally, the resulting $\mathbf{H}^{t}$ provides target-aware topological representations for anomaly detection.

\noindent\textbf{Semantic Pattern Encoding.}
In several scenarios, anomalies are mainly reflected in node attributes, such as rumors posted by users with normal structural patterns. To capture this complementary semantic information, \ourmethod further employs a semantic branch to detect attribute-level anomalies. As the first step, we unify heterogeneous node attributes across graphs through a landmark-based scattering mechanism. Specifically, given node features, PCA algorithm is used to project heterogeneous attributes into a unified dimension, denoted as $\mathbf{X}^{p}=\mathbf{X}\mathbf{W}_{p}$, where $\mathbf{W}_{p}$ denotes the PCA projection matrix. 
While this projection resolves dimensional heterogeneity, cross-graph distribution discrepancies may still remain. 
We therefore further apply landmark-based scattering to normalize the projected features across graphs and then employ an MLP for semantic feature encoding:
\begin{equation}
\mathbf{H}^{s}
=
\operatorname{MLP}\left(
\operatorname{Norm}\left(
{\mathbf{X}}^{p}\mathbf{X}_{l}^{\top}
\mathbf{K}^{-\frac{3}{2}}\mathbf{X}_{l}
\right)
\right),\qquad \
\mathbf{K}=\frac{\mathbf{X}_{l}\mathbf{X}_{l}^{\top}}
{|\mathcal{V}_{l}|-1},
\label{eq:semantic_scattering}
\end{equation}
where $\mathbf{X}_{l}$ is formed by randomly sampling a subset of nodes from $\mathbf{X}^{p}$ as landmarks, $\mathbf{K}$ captures their geometric structure, and $\mathbf{H}^{s}$ is the output semantic representations. The inverse spectral transformation rescales different feature directions according to the landmark geometry, yielding a more isotropic distribution in the latent space and reducing cross-graph feature discrepancies.

\noindent\textbf{Anomaly Evidence Coordination.}
The topological and semantic branches provide complementary anomaly evidence. As their importance may vary across nodes, we calculate anomaly scores at each branch separately and adaptively fuse their contributions.
Given the normal and anomalous support sets $\mathcal{S}_n$ and $\mathcal{S}_a$, we compute the branch-wise anomaly evidence as

\begin{equation}
q_{ij}^{b}
=\frac{(\mathbf{h}_{i}^{b})^{\top}\mathbf{h}_{j}^{b}}
{\|\mathbf{h}_{i}^{b}\|_2\|\mathbf{h}_{j}^{b}\|_2},
\qquad
e_i^{b}
=\log
\frac{
\sum_{v_j\in\mathcal{S}_a}\exp(\rho_b q_{ij}^{b})
}{
\sum_{v_j\in\mathcal{S}_n}\exp(\rho_b q_{ij}^{b})
},
\label{eq:stream_score}
\end{equation}

where $b\in\{t,s\}$ denotes the topological or semantic branch, $q_{ij}^{b}$ denotes the cosine similarity between query node $v_i$ and support node $v_j$, and $\rho_b$ is a learnable scaling factor. The key design principle here is that a query node is anomalous if it is more similar to anomalous supports than to normal ones.

Then, to account for the varying reliance of each node on topological and semantic signals, we further design a sample-level gating network. Since the topological descriptor $\mathbf{r}_i$ characterizes the behavior pattern of each node, we use it as the gating signal to estimate the relative importance of the two branches. The resulting weights are then applied to the corresponding branch-specific anomaly scores to obtain the final prediction:
\begin{equation}
[w_i^{t},w_i^{s}]
=\operatorname{Softmax}\left(\psi(\mathbf{r}_i)\right),
\qquad
\hat{y}_i
=\sigma\left(
w_i^{t}e_i^{t}+w_i^{s}e_i^{s}
\right),
\label{eq:adaptive_scoring}
\end{equation}
where $\psi(\cdot)$ is an MLP-based gating network, and $w_i^{t}$ and $w_i^{s}$ are the corresponding branch weights. %

By jointly modeling topological and semantic information, \ourmethod captures diverse anomaly patterns and adaptively balances the two signals at the sample level. Its enhanced expressiveness enables it to learn richer and more transferable knowledge of GAD from large-scale synthetic data. %

\subsection{Curriculum Learning from Synthetic Data}\label{Curriculum}

Although \oursyn provides abundant synthetic training data, the generated graph corpus contains graphs with diverse anomaly patterns and varying detection difficulties. In this case, treating all graphs equally to train \ourmethod may lead to unstable learning, especially when challenging patterns are underemphasized. To exploit large-scale synthetic data for GAD, we introduce a curriculum learning strategy that progresses from easy to hard, guiding the model to first capture clear anomaly patterns and then refine its discrimination ability on more challenging graphs.

\textbf{Staged Difficulty Progression.}
We use the controllable difficulty factor $\delta$ to organize synthetic graphs by difficulty. Training starts from graphs with more distinguishable anomalies, while harder graphs are progressively added to the training pool $\mathcal{D}^{(t)}$ at the $t$-th iteration:
\begin{equation}
\mathcal{D}^{(t)}
=\mathcal{G}_{i}^{s}\in\mathcal{D}_{\mathrm{syn}}
\mid
\delta_g\leq\Gamma(t)
,
\label{eq:curriculum}
\end{equation}
where $\Gamma(t)$ increases with training, progressively exposing the model to more challenging graphs and providing a smoother optimization trajectory.

\textbf{Dynamic Loss-Aware Soft Sampling.}
While $\delta$ is a predefined difficulty factor, it only reflects the intended difficulty during data generation and may not fully represent how difficult each graph is for the current model. Therefore, we further adjust graph sampling according to its learning status. For each synthetic graph, we maintain an exponential moving average (EMA) of its anomaly loss:
\begin{equation}
\bar{\mathcal{L}}_{i}^{(t)}
=(1-\beta)\bar{\mathcal{L}}_{i}^{(t-1)}
+
\beta \mathcal{L}_{i}^{(t)}(\hat{\mathbf{y}},\mathbf{y}),
\label{eq:loss_aware}
\end{equation}
where $\mathcal{L}_{i}^{(t)}$ is the average anomalous-node loss on $\mathcal{G}_{i}^{s}$ and $\beta$ is the momentum factor. Graphs with higher EMA losses receive larger sampling probabilities, while easier graphs are sampled less often, shifting training toward insufficiently learned patterns without discarding learned ones.

\textbf{Node-Level Focal Optimization.}
At a finer granularity, detection difficulty also varies across nodes within the same graph. To prevent easy samples from dominating optimization, we employ binary focal loss to emphasize poorly detected samples:
\begin{equation}
\mathcal{L}(\hat{{\mathbf{y}}},\mathbf{y})
=-\frac{1}{|\mathcal{V}|}
\sum_{i\in\mathcal{V}}
\omega_i
(1-p_i)^{\gamma}
\log p_i,
\label{eq:focal_loss}
\end{equation}
where $p_i=y_i \hat{y}_i+(1-y_i)(1-\hat{y}_i)$ denotes the predicted probability of node $v_i$ belonging to its ground-truth class, the term $(-\log p_i)$ corresponds to the standard binary cross-entropy loss, and $\omega_i$ is a weighting factor that emphasizes anomalous nodes. To further emphasize poorly detected samples, we introduce the focusing factor $(1-p_i)^{\gamma}$ to reduce the influence of easy samples and assign greater importance to difficult ones, with $\gamma$ controlling the degree of emphasis.

Furthermore, to explicitly preserve the discriminative ability of both branches, we jointly supervise the fused prediction and the individual topological and semantic predictions:
\begin{equation}
\mathcal{L}
=\mathcal{L}(\hat{\mathbf{y}},\mathbf{y})
+
\mathcal{L}(\hat{\mathbf{y}}^{t},\mathbf{y})
+
\mathcal{L}(\hat{\mathbf{y}}^{s},\mathbf{y}),
\label{eq:overall_loss}
\end{equation}
where $\hat{\mathbf{y}}^{t}=\sigma(\mathbf{e}^{t})$ and $\hat{\mathbf{y}}^{s}=\sigma(\mathbf{e}^{s})$ denote the predictions from the topological/semantic branches.

Overall, the proposed curriculum coordinates learning at both graph and sample levels. The staged progression controls when graphs of different predefined difficulties are introduced, the loss-aware sampling dynamically emphasizes graphs that remain challenging for the current model, and the focal objective further focuses learning on difficult nodes within each graph. Together, these designs enable the model to better leverage the diverse anomaly knowledge contained in the synthetic corpus.

%% file: Section/experiment.tex
\subsection{Experimental Setup}

\textbf{Datasets.}
To evaluate the detection capability of \ourmethod trained on synthetic graphs by \oursyn, we conduct comparative experiments on 14 real-world datasets following~\cite{liu2026rethinking}, including five citation networks (\textit{Cora}, \textit{Pubmed}, \textit{Citeseer}, \textit{CS}, \textit{ACM}), three social networks (\textit{BlogCatalog}, \textit{Flickr}, \textit{Facebook}), three e-commerce networks (\textit{Amazon}, \textit{Photo}, \textit{YelpChi}), and three behavioral networks (\textit{Weibo}, \textit{Questions}, \textit{Reddit}). Appendix~\ref{appe:data} provides detailed dataset statistics.

\textbf{Baselines.}
We compare \oursyn$+$\ourmethod on various baselines, including two GNNs (GCN~\cite{kipf2016semi} and GAT~\cite{velivckovic2017graph}), three unsupervised GAD methods (CoLA~\cite{liu2021anomaly}, SmoothGNN~\cite{dong2025smoothgnn}, and ANEMONE~\cite{jin2021anemone}), five supervised GAD methods (AHFAN~\cite{wang2025graph}, BWGNN~\cite{tang2022rethinking}, BGNN~\cite{ivanov2021boost}, GHRN~\cite{gao2023addressing}, and GGAD~\cite{qiao2024generative}), and four generalist GAD methods (ARC~\cite{liu2024arc}, UNPrompt~\cite{niu2024zero}, IA-GGAD~\cite{zhang2026ia}, and ReFi-GAD~\cite{liu2026rethinking}). Details of baselines are provided in Appendix~\ref{appe:baseline}.

\textbf{Evaluation and Implementation.}
\ourmethod is exclusively trained on the graphs generated by \oursyn and evaluated on real-world datasets. For existing generalist GAD methods, we train them on real-world source graphs to fully realize their performance. Specifically, following previous work~\cite{liu2026rethinking}, we divide the 14 datasets into two disjoint and domain-balanced groups, train the models on one group and test on another, and then swap the two groups. We also train these generalist models on data generated by \oursyn to verify the quality of synthetic data.  We use AUROC and AUPRC as the evaluation metrics and report the mean and standard deviation over five runs. For our method, the default value of $K$ is 30, and the support set size is kept consistent across all few-shot generalist methods. All methods use consistent hyperparameters across target datasets without parameter updates during testing.

\input{tables/main_table}

\subsection{Experimental Results}

\noindent\textbf{Performance Comparison.}
We compare \ourmethod with baselines on 14 real-world datasets. Generalist GAD methods are trained on real-world data or data synthesized by \oursyn. Table~\ref{tab:main} reports the average AUROC results, with additional results provided in Appendix~\ref{appe:main_exp}. We have the following observations. 
\ding{182}~\oursyn$+$\ourmethod achieves the best performance on all 14 datasets and the highest average AUROC, highlighting the effectiveness of jointly leveraging large-scale synthetic data, a stronger model architecture, and curriculum learning. \ding{183}~\oursyn enables existing generalist GAD methods to achieve performance comparable to training on real-world data, demonstrating the viability of synthetic data for generalist GAD.
\ding{184} \ourmethod consistently outperforms existing generalist GAD methods by a large margin, e.g., surpassing ReFi-GAD by 5.39\%. This demonstrates that synthetic graphs by \oursyn can provide sufficient transferable anomaly knowledge, while the architecture design of \ourmethod further unlocks this capability. 

\begin{wrapfigure}{r}{0.35\textwidth}
  \vspace{-4mm}
  \includegraphics[width=0.35\textwidth]{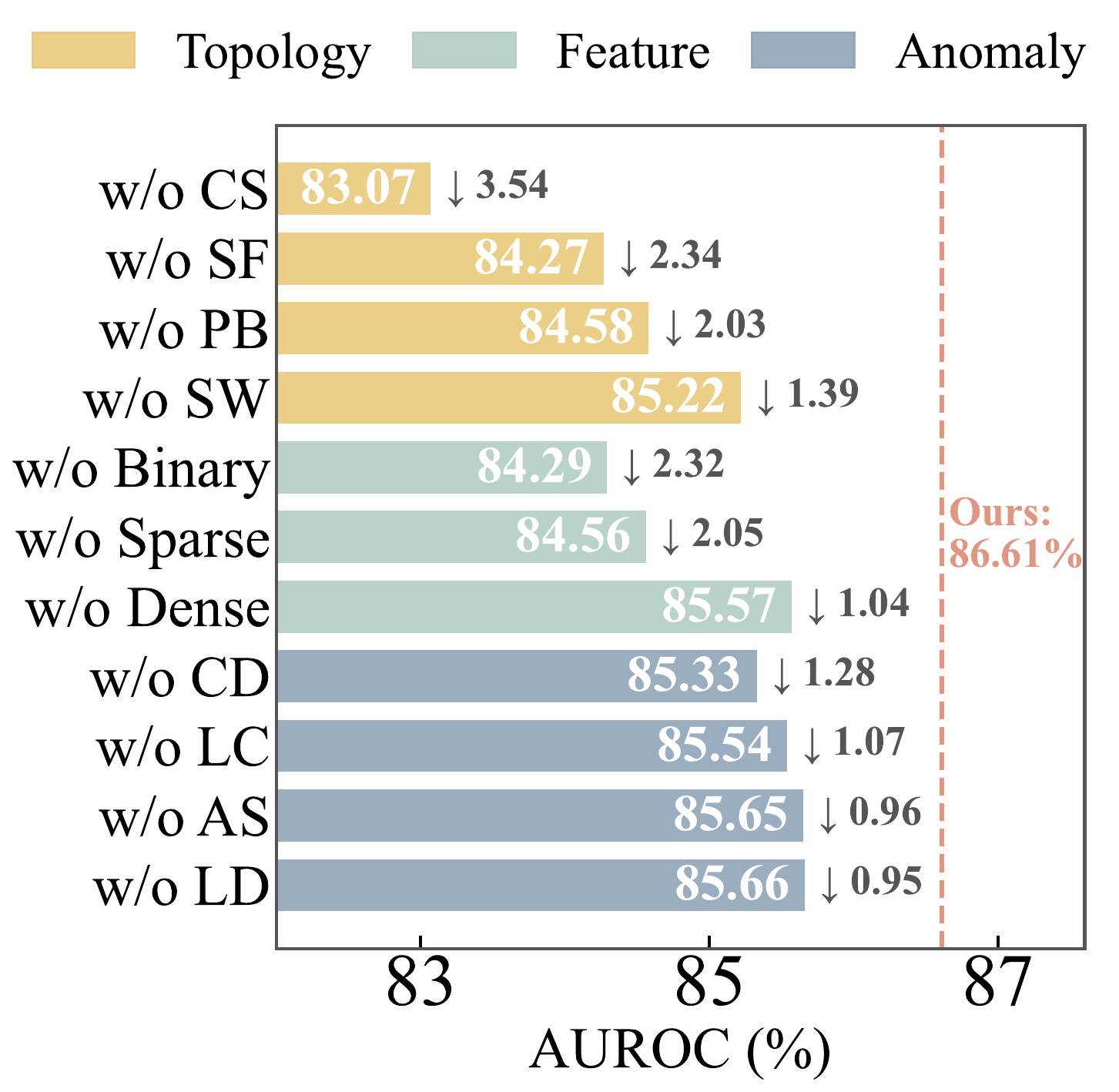}
    \vspace{-6mm}
  \caption{Ablation for \oursyn.}
  \label{fig.ablation_data}
    \vspace{-5mm}
\end{wrapfigure}

\noindent\textbf{Ablation Study of \oursyn.}
To evaluate the contribution of different strategies in \oursyn, we individually remove each structure, feature, or anomaly generation strategy. As shown in Fig.~\ref{fig.ablation_data}, all variants consistently degrade performance, indicating that these strategies provide complementary diversity for synthetic training. In particular, removing one structure generation strategy causes an average 2.33\% drop, highlighting the importance of diverse topology patterns for cross-domain generalization. Removing feature and anomaly generation strategies also leads to average declines of 1.80\% and 1.07\%, respectively, confirming that diverse semantic contexts and anomaly patterns are both important for learning transferable anomaly knowledge. Detailed results are provided in Appendix~\ref{appe:ablation_data}.

\begin{wrapfigure}{r}{0.35\textwidth}
  \vspace{-3mm}
  \includegraphics[width=0.35\textwidth]{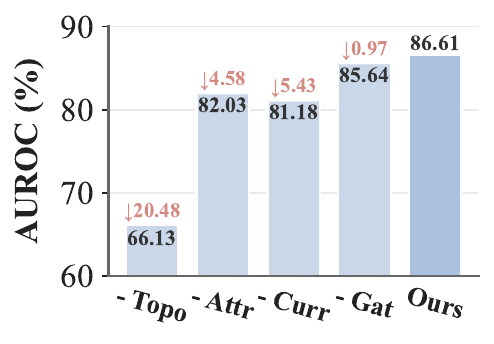}
    \vspace{-6mm}
  \caption{Ablation for \ourmethod.}
  \label{fig.ablation_model}
    \vspace{-3mm}
\end{wrapfigure}
\noindent\textbf{Ablation Study of \ourmethod.}
To assess the contribution of each component in \ourmethod, we consider four variants: removing the topological stream (\textbf{- Topo}), removing the semantic stream (\textbf{- Attr}), removing curriculum learning (\textbf{- Curr}), and replacing target-aware gating with naive fusion (\textbf{- Gat}). As shown in Fig.~\ref{fig.ablation_model}, all variants degrade performance. Removing the topological stream causes the largest drop of 20.48\%, demonstrating the importance of structural anomaly signals. Removing the semantic stream and curriculum learning decreases performance by 4.58\% and 5.43\%, respectively. These results confirm that all designs work complementarily to support the generalist anomaly detection capability of \ourmethod. Detailed results are provided in Appendix~\ref{appe:ablation_model}.

\begin{wrapfigure}{r}{0.3\textwidth}
  \vspace{-5mm}
  \includegraphics[width=0.3\textwidth, trim = 0 0 0 55, clip]{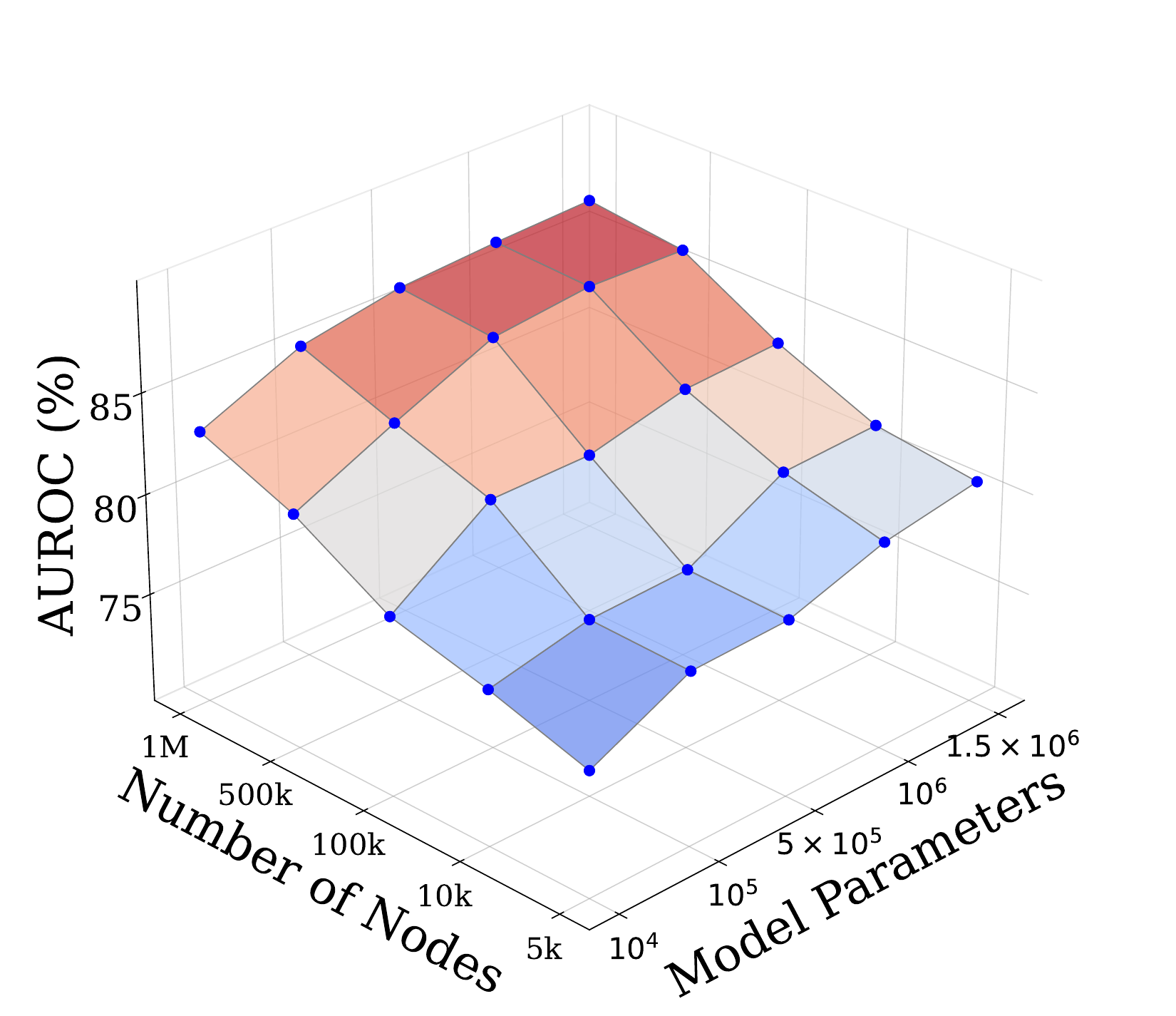}
    \vspace{-6mm}
  \caption{Scaling analysis.}
  \label{fig.scal}
    \vspace{-3mm}
\end{wrapfigure}

\noindent\textbf{Scaling Analysis.}
To examine the scaling capability of \ourmethod, we vary the training data size from 5k to 1M nodes and the model parameters from $10^4$ to $1.5\times10^6$, as shown in Fig.~\ref{fig.scal}. 
The performance of \ourmethod improves as both training data and model capacity increase. Notably, models with more than one million parameters require larger training sets (e.g., 500k nodes) to fully exploit their capacity, and the best performance is achieved in the ``large data--large model'' region, where AUROC exceeds 86\%. These results reveal a clear scaling trend and highlight the potential of \ourmethod for generalist GAD. Meanwhile, the proposed \oursyn provides the scalable training data needed to support model scaling.

\begin{wrapfigure}{r}{0.3\textwidth}
  \vspace{-8mm}
  \includegraphics[width=0.3\textwidth, trim = 0 0 0 0, clip]{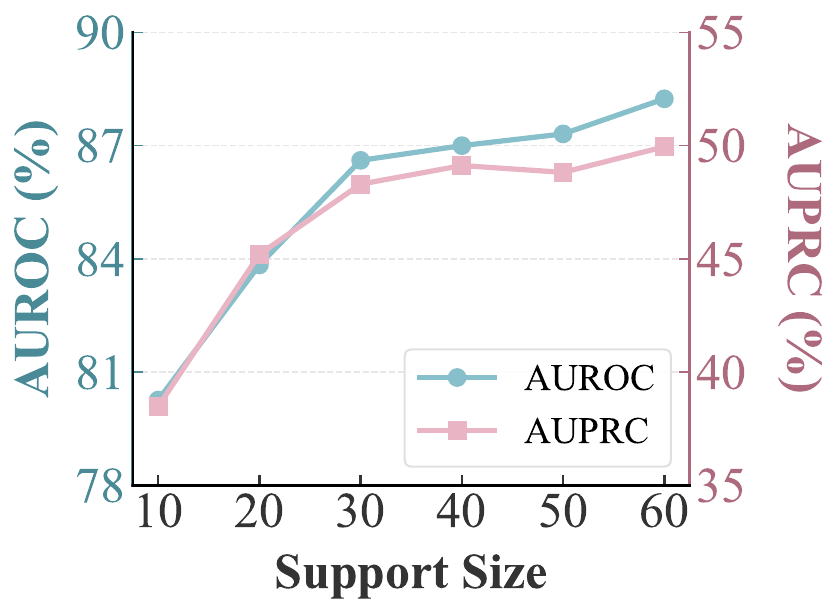}
    \vspace{-6mm}
  \caption{Sensitivity to $K$.}
  \label{fig.ps}
    \vspace{-3mm}
\end{wrapfigure}

\noindent\textbf{Parameter sensitivity experiment.}
We investigate the sensitivity of \ourmethod to the support size $K$ by varying $K$ within $\{10, 20, \dots, 60\}$. The experimental results are illustrated in Fig.~\ref{fig.ps}. As shown in the figure, the performance consistently improves with larger $K$, as more support samples provide richer target-domain context. However, the improvement becomes smaller when $K>30$, indicating that a moderate number of support samples is sufficient for effective target-domain adaptation. This also demonstrates that \ourmethod does not require a large support set at inference time. We additionally study the sensitivity to the focal parameter $\gamma$, with the corresponding results and analysis provided in Appendix~\ref{app:sensitivity}.

\begin{wrapfigure}{r}{0.28\textwidth}
  \vspace{-8mm}
  \includegraphics[width=0.28\textwidth, trim = 0 0 0 0, clip]{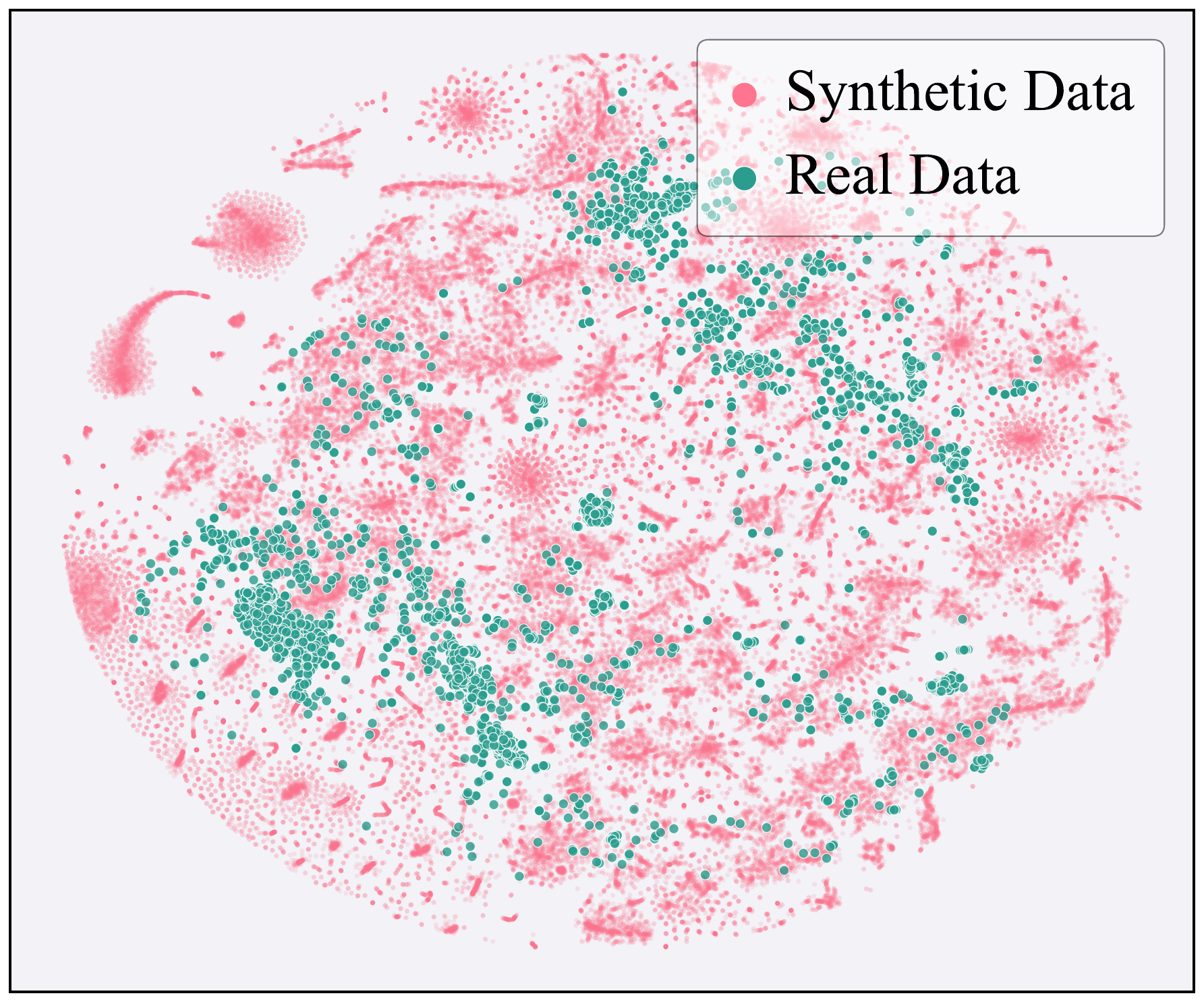}
    \vspace{-6mm}
  \caption{Visualization.}
  \label{fig.embed}
    \vspace{-3mm}
\end{wrapfigure}
\noindent\textbf{Visualization of Data Coverage.} 
To compare the coverage of real-world and synthetic data, we sample nodes from multiple real and generated graphs, align their embeddings with PCA, and visualize them using t-SNE. As shown in Fig.~\ref{fig.embed}, real-world samples occupy several isolated regions, whereas synthetic samples form a broader and more continuous distribution that significantly overlaps with real data while covering sparsely represented regions. This indicates that \oursyn provides richer and more continuous training coverage beyond limited real-world graphs. See Appendix~\ref{app:visualization} for more visualizations.

%% file: tables/main_table.tex
\begin{table*}[!t]
  \centering
  \caption{Detection performance in terms of AUROC (\%). Highlighted are the results ranked \first{first}, \second{second}, and \third{third}. \colorbox{agforgePurple}{Purple-shaded rows} denote models trained with \oursyn-generated data.}
  \normalsize
  
  \setlength{\tabcolsep}{3pt} 
  \newcolumntype{Y}{>{\centering\arraybackslash}p{1cm}}
    \renewcommand{\arraystretch}{1.2}
    
  \resizebox{\textwidth}{!}{
    \begin{tabular}{l|*{14}{Y}|c}
    \toprule
    {Method} 
          & Cite & CS    & ACM   & Blog & Amz & Photo & Weibo & Cora  & Pubmed & Flickr & FB & Yelp & Quest & Reddit & {Rank} \\
          \midrule
    \rowcolor[rgb]{ .906,  .902,  .902} \multicolumn{16}{c}{GAD Methods} \\
    GCN   & 48.32  & 56.19  & 50.43  & 43.06  & 58.69  & 47.70  & 46.40  & 32.41  & 33.68  & 38.07  & 76.35  & 51.21  & 41.37  & 46.80  & 16.07  \\
    GAT   & 63.11  & 59.25  & 61.48  & 60.50  & 54.84  & 45.63  & 73.51  & 56.07  & 67.21  & 56.05  & 55.33  & 50.43  & 57.37  & 43.39  & 14.64  \\
    CoLA  & 73.81  & 65.99  & 54.94  & 60.58  & 65.02  & 65.18  & 41.08  & 66.02  & 70.09  & 60.64  & 70.88  & 52.41  & 53.23  & 50.60  & 12.43  \\
    SmoothGNN & 90.42  & 78.65  & 78.00  & 73.06  & 50.75  & 44.80  & 85.96  & 88.39  & 78.22  & 76.95  & 53.62  & 61.30  & 59.01  & 54.09  & 8.64  \\
    ANEMONE & 41.36  & 42.18  & 45.53  & 35.18  & 43.51  & 49.22  & 35.57  & 47.08  & 37.65  & 37.07  & 45.55  & 50.88  & 51.55  & 52.41  & 17.29  \\
    AHFAN & 48.02  & 59.09  & 63.70  & 59.60  & 29.50  & 52.16  & 63.03  & 55.89  & 64.08  & 62.77  & 32.11  & 49.76  & 54.01  & 46.90  & 15.64  \\
    BWGNN & 60.91  & 59.12  & 61.80  & 67.26  & 50.40  & 60.82  & 81.62  & 63.62  & 68.19  & 68.37  & 41.55  & 52.76  & 57.59  & \second{{60.24}} & 11.64  \\
    BGNN  & 55.14  & 52.14  & 52.01  & 52.65  & 52.17  & 47.26  & 77.47  & 52.66  & 58.46  & 52.56  & 65.80  & 50.67  & 56.80  & 41.80  & 15.57  \\
    GHRN  & 74.25  & 78.01  & 76.88  & 65.09  & 52.93  & 51.99  & 57.77  & 68.18  & 75.15  & 64.32  & 33.07  & 51.72  & \second{{59.81}} & 53.45  & 11.29  \\
    GGAD  & 67.03  & 64.76  & 67.02  & 58.83  & 50.36  & 58.72  & 70.23  & 71.15  & 69.82  & 59.79  & 20.24  & 52.14  & 58.92  & 53.78  & 12.79  \\
    \midrule
    \rowcolor[rgb]{ .906,  .902,  .902} \multicolumn{16}{c}{Generalist GAD Methods} \\
    ARC   & 89.96  & 82.73  & 79.83  & 73.98  & \second{{79.77}} & 75.54  & 88.05  & 83.59  & 84.44  & 75.61  & 61.11  & 54.47  & \third{{59.48}} & 58.94  & 5.86  \\
    \rowcolor[rgb]{0.93, 0.90, 0.97} +\oursyn   & 89.70  & 82.99  & 79.94  & 74.16  & 71.23  & \third{{75.67}} & 87.70  & 85.19  & 85.21  & 74.48  & 72.09  & 50.56  & 58.70  & \third{{59.40}} & 6.57  \\
    \midrule
    UNPrompt & 72.59  & 74.38  & 73.99  & 69.02  & 63.14  & 69.61  & 49.19  & 71.23  & 82.48  & 69.68  & 76.34  & 52.36  & 46.17  & 56.14  & 10.36  \\
    \rowcolor[rgb]{0.93, 0.90, 0.97} +\oursyn   & 76.82  & 76.11  & 74.55  & 69.18  & 62.98  & 71.06  & 49.37  & 67.47  & 81.78  & 69.64  & 74.48  & 50.63  & 45.27  & 56.24  & 10.71  \\
    \midrule
    IA-GGAD & 91.84  & 93.04  & 91.19  & \third{{74.76}} & \third{{71.93}} & 70.23  & \third{{91.10}} & 87.90  & 87.84  & 69.05  & 79.86  & 52.38  & 57.82  & 57.97  & 5.43  \\
    \rowcolor[rgb]{0.93, 0.90, 0.97} +\oursyn   & 92.18  & 91.49  & 90.14  & \second{{74.81}} & 70.87  & 72.50  & \second{{91.55}} & 87.83  & 88.56  & 67.04  & 80.70  & 51.38  & 58.22  & 57.84  & 5.71  \\
    \midrule
    ReFi-GAD & \third{{95.60}} & \third{{98.69}} & \third{{93.94}} & 69.96  & 65.98  & 73.78  & 88.52  & \second{{95.74}} & \second{{97.07}} & \second{{81.25}} & \second{{92.79}} & \third{{68.90}} & 57.47  & 57.40  & 4.50  \\
    \rowcolor[rgb]{0.93, 0.90, 0.97} +\oursyn   & \second{{96.40}} & \second{{99.03}} & \second{{93.96}} & 69.42  & 68.32  & \second{{76.58}} & 89.02  & \third{{94.56}} & \third{{96.66}} & \third{{77.28}} & \third{{90.80}} & \second{{73.65}} & 57.78  & 58.74  & 3.86  \\
    \midrule
    \rowcolor[rgb]{0.93, 0.90, 0.97} \ourmethod & \first{{96.52}} & \first{{99.43}} & \first{{96.24}} & \first{{76.19}} & \first{{87.10}} & \first{{82.15}} & \first{{93.53}} & \first{{96.41}} & \first{{97.77}} & \first{{86.46}} & \first{{97.41}} & \first{{74.46}} & \first{{64.62}} & \first{{64.31}} & \first{{1.00}} \\
    \bottomrule
    \end{tabular}}%
  \label{tab:main}%
\end{table*}

%% file: Section/conclusion.tex
In this paper, we explore the feasibility of synthetic data-based training for generalist GAD. We propose \oursyn, which synthesizes diverse anomalous graphs through a three-stage cascaded ``structure-feature-anomaly'' pipeline. To fully exploit the large-scale synthetic data, we further develop \ourmethod, a topology-semantic coordinated generalist GAD model with greater capacity and expressiveness, together with a curriculum learning strategy for effective training on graphs with varying detection difficulties. Experiments on 14 real-world datasets show that \ourmethod trained solely on synthetic data achieves strong detection, cross-domain generalization, and scaling capability, highlighting the potential of fully synthetic training for generalist GAD. The main \textbf{limitation} is the lack of systematic quality evaluation of synthetic data. Therefore, a promising \textbf{future direction} is to develop effective criteria for data evaluation and selection.

%% file: Section/appendix.tex
\section{Related Works}\label{appe:related_work}

\subsection{Graph Anomaly Detection} 
Graph Anomaly Detection (GAD) aims to identify anomalous nodes, edges, or subgraphs that deviate significantly from normal graph patterns. Most traditional GAD methods follow a dataset-specific paradigm, requiring training on each target graph before anomaly detection. Early studies commonly adopt graph neural networks (GNNs), such as GCN~\cite{kipf2016semi} and GAT~\cite{velivckovic2017graph}, as basic encoders and optimize reconstruction or classification objectives to distinguish anomalous nodes from normal ones.

To improve anomaly discrimination, subsequent studies have explored different learning paradigms tailored to GAD. CoLA~\cite{liu2021anomaly} develops a contrastive self-supervised framework that measures the agreement between each node and its sampled local subgraph, while ANEMONE~\cite{jin2021anemone} extends this idea to multi-scale contrastive learning at both patch and context levels. From the spectral perspective, BWGNN~\cite{tang2022rethinking} reveals the frequency ``right-shift'' caused by anomalous nodes and introduces beta-wavelet filters to capture anomaly-related spectral patterns. More recently, SmoothGNN~\cite{dong2025smoothgnn} exploits the different smoothing behaviors of normal and anomalous nodes for unsupervised detection.

Despite their promising performance, these methods rely on dataset-specific training, requiring retraining for each target graph and introducing substantial deployment and maintenance costs, which limits their practical scalability.

\subsection{Generalist Graph Anomaly Detection}

To reduce the repeated training required by dataset-specific GAD, recent studies have begun to explore Generalist Graph Anomaly Detection (GGAD), which aims to learn a unified detector that can generalize to unseen graphs without graph-specific retraining or fine-tuning. ARC~\cite{liu2024arc} pioneers this direction by introducing in-context learning, using a small number of target-domain normal samples to support anomaly detection on unseen graphs. UNPrompt~\cite{niu2024zero} further explores zero-shot GGAD through unified neighborhood prompts. AnomalyGFM~\cite{qiao2025anomalygfm} develops a GAD-oriented graph foundation model by learning graph-agnostic normal and abnormal prototypes, supporting both zero-shot inference and few-shot adaptation.

More recent studies further improve cross-domain transfer from different perspectives. IA-GGAD~\cite{zhang2026ia} addresses feature and structural shifts across graphs through invariant representation and affinity learning. ReFi-GAD~\cite{liu2026rethinking} introduces relational fingerprints to reduce cross-domain feature heterogeneity and combines generalizable representation learning with target-aware refinement. These advances demonstrate the potential of GGAD for transferable anomaly detection. Despite these advances, existing GGAD methods are constrained by limited training data, making it difficult to further scale up training and improve model capacity and generalization.

\subsection{Synthetic Data for Foundation Models}

Synthetic data have recently shown strong potential for training foundation models without relying on large-scale real-world corpora. TabPFN~\cite{hollmann2025accurate} is a representative example, which learns a general tabular prediction algorithm by pretraining on millions of synthetic datasets and directly performs in-context prediction on unseen tables. TabICL~\cite{qu2025tabicl} further scales this paradigm to substantially larger tabular datasets through a more efficient in-context learning architecture. Beyond tabular data, TimePFN~\cite{taga2025timepfn} generates multivariate time series from diverse Gaussian-process priors for zero- and few-shot forecasting, while RDB-PFN~\cite{wang2026relational} constructs synthetic relational databases with structural priors to learn transferable relational prediction capabilities. These studies demonstrate that carefully designed synthetic data can provide scalable supervision for foundation model training across different data modalities.

For graph data, synthetic training is still at an early stage. GraphPFN~\cite{eremeev2025graphpfn} explores pretraining on fully synthetic graphs, but its generated topology and feature patterns remain relatively limited. More importantly, to the best of our knowledge, synthetic anomalous graph generation for generalist GAD remains largely unexplored.

\section{Details of \oursyn} \label{app:oursyn}
In this paper, to alleviate the data drought in generalist GAD, we design \textbf{\oursyn}, an \textbf{A}nomalous \textbf{G}raph generation \textbf{Forge} for automatically synthesizing diverse anomalous graphs at scale. Given the strong heterogeneity of real-world graphs, our goal is not to approximate any particular real-world graph distribution, but to construct a synthetic space spanning diverse graph environments and anomaly patterns. To this end, we decouple the generation process into a three-stage cascaded pipeline: \textcolor{challengeColor}{\textbf{graph structure generation} \ding{220} \textbf{node feature generation} \ding{220} \textbf{anomaly injection}}.
Let $\tau$, $\phi$, and $\alpha$ denote the structure, feature, and anomaly types, respectively, while $\delta$ controls the anomaly difficulty. The overall generation process is formulated as
\begin{equation}
\begin{cases}
\mathbf{A}_{0}^{s}
= g_{\mathrm{str}}(N;\tau),
& \text{Step 1: Structure Generation}, \\

\mathbf{X}_{0}^{s}
= g_{\mathrm{feat}}(\mathbf{A}_{0}^{s},D;\phi),
& \text{Step 2: Feature Generation}, \\

(\mathbf{A}^{s},\mathbf{X}^{s},\mathbf{Y}^{s})
= g_{\mathrm{ano}}(\mathbf{A}_{0}^{s},\mathbf{X}_{0}^{s};\alpha,\delta),
& \text{Step 3: Anomaly Injection},
\end{cases}
\label{eq:syn_generation}
\end{equation}
where $N$ and $D$ denote the number of nodes and feature dimensionality, respectively, $\mathbf{A}_{0}^{s}$ and $\mathbf{X}_{0}^{s}$ represent the graph topology and node attributes before anomaly injection, while $\mathbf{A}^{s}$, $\mathbf{X}^{s}$, and $\mathbf{Y}^{s}$ denote the final anomalous topology, node attributes, and anomaly labels. By varying these generation factors, we construct diverse graphs varying in scale, structure, feature patterns, and anomaly patterns.

\noindent\textbf{Step 1:~Structure Generation.}
To simulate the structural diversity of real-world graphs, we design four topology generation mechanisms with distinct structural characteristics. Given the randomly sampled graph size $N$ and structure type $\tau$, the base graph structure $\mathbf{A}_{0}^{s}$ is generated as
\begin{equation}
N \sim \mathcal{U}_{\mathbb{Z}}(N_{\min},N_{\max}), \qquad
\tau \sim \mathcal{U}\{\mathrm{CS},\mathrm{SF},\mathrm{PB},\mathrm{SW}\}, \qquad
\mathbf{A}_{0}^{s}=g_{{str}}(N;\tau).
\end{equation}
where $N$ is uniformly sampled from $[N_{\min}, N_{\max}]$, $\tau$ denotes the sampled structure type, and $\mathrm{CS}$, $\mathrm{SF}$, $\mathrm{BP}$, and $\mathrm{SW}$ denote community-structured, scale-free, projected bipartite, and small-world graphs, respectively. The characteristics of these structure models are as follows:

\textbf{Community-structured graph (CS).}
This type is based on the stochastic block model (SBM), forming explicit communities where intra-community connections are denser than inter-community ones. Such community patterns are common in many real-world graphs, including social and recommendation networks. Specifically, we partition the $N$ nodes into $C$ communities and randomly establish edges according to
\begin{equation}
P(a_{ij}^{s} = 1)=
\begin{cases}
p_{\mathrm{in}}, & c_i=c_j,\\
p_{\mathrm{out}}, & c_i\neq c_j,
\end{cases}
\qquad
p_{\mathrm{in}}>>p_{\mathrm{out}},
\label{eq:sbm}
\end{equation}
where $c_i$ denotes the community assignment of node $v_i$. The community sizes, $p_{\mathrm{in}}$, and $p_{\mathrm{out}}$ are varied across generated graphs to produce different community structures.

\textbf{Scale-free graph (SF).}
This type simulates hub-dominated structures commonly observed in real-world graphs such as citation and communication networks, where a small fraction of nodes have more connections than others. The graph is generated through a sequential growth process, where each new node $v_i$ establishes $m$ edges with existing nodes. Under preferential attachment, an existing node $v_j$ is selected with probability
\begin{equation}
P(a_{ij}^{s} = 1) = \frac{d_j^{i-1}}{\sum_{v_k\in\mathcal{V}_{i-1}} d_k^{i-1}},
\label{eq:sf}
\end{equation}
where $\mathcal{V}_{i-1}$ denotes the node set before adding $v_i$, and $d_j^{(i-1)}$ is the current degree of $v_j$. A triad-formation step is further performed with probability $p_t$ to introduce local clustering. By varying $m$ and $p_t$, we generate scale-free graphs with different connectivity densities and clustering patterns.

\textbf{Projected bipartite graph (PB).}
This type simulates homogeneous structures induced from heterogeneous interactions, such as user-user relations formed through shared items in user-item networks. We first construct a sparse bipartite interaction matrix $\mathbf{B}\in{0,1}^{N\times M}$ between $N$ target nodes and $M$ auxiliary nodes. The bipartite graph is then projected onto the target node set:
\begin{equation}
a_{ij}^{s} = \mathbb{I}\left(
\sum_{r=1}^{M} B_{ir}B_{jr}>0
\right),
\quad i\neq j,
\label{eq:bp}
\end{equation}
where two target nodes are connected if they share at least one auxiliary neighbor. By varying $M$ and interaction density, we generate projected graphs with diverse connectivity patterns.

\textbf{Small-world graph (SW).}
This type simulates small-world structures with strong local connectivity and short global paths, as commonly observed in social and communication networks. We first arrange $N$ nodes on a ring and connect each node to $k$ surrounding nodes along the ring, forming a regular local structure:
\begin{equation}
a_{ij}^{s} = \mathbb{I}\left(
0<\ell_{\mathrm{ring}}(i,j)\leq \frac{k}{2}
\right),
\quad
\ell_{\mathrm{ring}}(i,j) = \min(|i-j|,N-|i-j|).
\label{eq:sw}
\end{equation}
where $\ell_{\mathrm{ring}}(i,j)$ denotes the shortest distance between $v_i$ and $v_j$ on the ring. Each edge is then rewired with probability $p_r$ to introduce long-range connections.

These four mechanisms produce the base topology $\mathbf{A}_{0}^{s}$ with diverse structural characteristics. To guide subsequent feature generation, each node is assigned a group label $c_i$. For CS graphs, $c_i$ follows the SBM block assignment; for other types of graphs, groups are obtained either by Louvain community detection or random partitioning, simulating community patterns in real-world graphs.

\noindent\textbf{Step 2:~Feature Generation.}
Apart from topology, real-world graphs also exhibit heterogeneity in node attributes. Accordingly, we construct three types of node attributes: dense continuous, sparse high-dimensional, and binary features. Given the base topology $\mathbf{A}_{0}^{s}$, we first construct base features according to the node groups $c$ and then transform them into different feature types. Formally,
\begin{equation}
D \sim \mathcal{U}_{\mathbb{Z}}(D_{\min},D_{\max}), \qquad
\phi \sim \mathcal{U}\{\mathrm{Dense},\mathrm{Sparse},\mathrm{Binary}\}, \qquad
\mathbf{X}_{0}^{s}=g_{\mathrm{feat}}(\mathbf{A}_{0}^{s},D;\phi).
\end{equation}
\textbf{\textit{Step 2.1:~Base feature generation.}}
We first assign each group $c$ a low-dimensional latent center $\boldsymbol{\mu}_c\in\mathbb{R}^{d_z}$. For each node $v_i$, we sample a small node-specific variation $\boldsymbol{\xi}_i$ from a shared mixture distribution and add it to the corresponding group center before projecting the resulting representation to the target feature dimension:
\begin{equation}
\begin{cases}
    \mathbf{z}_i = \boldsymbol{\mu}_{c_i} + \boldsymbol{\xi}_i\\
    \overline{\mathbf{x}}^s_{i} = \mathbf{z}_i\mathbf{W} + \boldsymbol{\epsilon}_i,
\end{cases}
\label{eq:group_feature}
\end{equation}
where $\mathbf{W}\in\mathbb{R}^{d_z\times D}$ is a random projection matrix and $\boldsymbol{\epsilon}_i$ is Gaussian noise. This design preserves group-level similarity and node-specific variation while introducing dependencies across feature dimensions, better reflecting real-world feature distributions.

\textbf{\textit{Step 2.2:~Feature-type transformation.}}
Based on the generated base feature $\mathbf{x}_i^s$, we further construct three feature types to simulate diverse node attributes in real-world graphs. Specifically, the dense feature directly retains the base representation, the sparse feature applies random sparsification, and the binary feature converts each dimension into a binary value:
\begin{equation}
\mathbf{x}_{i}^{s}
=\begin{cases}
\overline{\mathbf{x}}^s_{i},
& \phi=\mathrm{Dense},\\
\mathrm{ReLU}(\overline{\mathbf{x}}^s_{i})\odot\mathbf{m}_i,
& \phi=\mathrm{Sparse},\\
\mathbb{I}\left(
\sigma(\overline{\mathbf{x}}_{i}^{s})>\boldsymbol{\tau}_i
\right), \quad 
\boldsymbol{\tau}_i\sim U(0,1),
& \phi=\mathrm{Binary},
\end{cases}
\label{eq:feature_type}
\end{equation}
where $\mathbf{m}_i$ is a random binary mask controlling sparsity, $\sigma(\cdot)$ denotes the sigmoid function, and $\boldsymbol{\tau}_i$ is sampled from $U(0,1)$. By varying the latent distribution, projection, and feature type, we generate diverse feature patterns with different sparsity and statistics, enriching attribute environments for generalist GAD training.

\noindent\textbf{Anomaly Injection.}
After constructing the base structure and node features, we further inject anomalies with different patterns to simulate diverse abnormal behaviors in different graph scenarios. Given the anomaly type $\alpha$ and difficulty factor $\delta$, the injection process is formulated as
\begin{equation}
\alpha \sim \mathcal{U}\{\mathrm{LD},\mathrm{LC},\mathrm{AS},\mathrm{CD},\mathrm{Mix}\}, \quad
\delta \sim \mathcal{U}(\delta_{\min},\delta_{\max}), \quad
(\mathbf{A}^{s},\mathbf{X}^{s},\mathbf{Y}^{s})
= g_{\mathrm{ano}}(\mathbf{A}_{0}^{s},\mathbf{X}_{0}^{s};\alpha,\delta),
\end{equation}
where $\mathrm{LD}$, $\mathrm{LC}$, $\mathrm{AS}$, $\mathrm{CD}$ and $\mathrm{Mix}$ denote different anomaly injection strategies, specified as:

\textbf{Local Densification (LD).}
This type simulates collective anomalies that form unusually dense local groups. We first sample an anomalous node set $\mathcal{V}_a$ and partition it into local groups. For each anomalous group $\mathcal{V}_g^a$, we increase its internal connectivity and assign a shared feature pattern:
\begin{equation}
P(a_{ij}^{s}=1)\propto \delta,\quad 
\mathbf{x}_{i,j}^{s}=\boldsymbol{\mu}_{g}^{a}+\boldsymbol{\epsilon}_{i,j},
\quad v_i,v_j\in\mathcal{V}_{g}^{a},
\end{equation}
where $\boldsymbol{\mu}_{g}^{a}$ is a sampled feature center for the anomalous group. As $\delta$ increases, the anomalous nodes become more densely interconnected, making the group increasingly resemble a coherent normal community. This pattern can be observed in coordinated behaviors such as money-laundering rings or fake-order groups.

\textbf{Link Concentration (LC).}
This type simulates coordinated anomalies such as malicious review attacks and DDoS attacks, where a large number of anomalous nodes collectively interact with a small number of prominent targets despite exhibiting attributes incongruent with those targets. Specifically, we sample a target set $\mathcal{V}_t$ and redirect the anomalous nodes toward these targets:
\begin{equation}
a_{ij}^{s}=1,\qquad
\mathbf{x}_{i}^{s}
=-\lambda(\delta)\mathbf{\hat{x}}_{t}
+\boldsymbol{\epsilon}_{i},
\quad
v_i\in\mathcal{V}_{a},\ v_j\in\mathcal{V}_{t},
\end{equation}
where $\mathbf{\hat{x}}_{t}$ denotes the mean feature of the target nodes and $\lambda(\delta)$ controls the magnitude of the feature shift. As $\delta$ increases, $\lambda(\delta)$ decreases, weakening the feature discrepancy, while the probability of activating additional cross-community rewiring and local feature alignment increases, further obscuring the co-target structure and feature inconsistency.

\textbf{Attribute Suppression (AS).}
This type simulates anomalous entities that remain active in their interactions while revealing little attribute information, as may occur in scenarios such as shell companies. Specifically, we connect each anomalous node to a randomly sampled target set $\mathcal{V}^{r}$ while suppressing its attributes:
\begin{equation}
a_{ij}^{s}=1,\qquad \mathbf{x}_{i}^{s} = 0 + (\eta(\delta)\boldsymbol{\epsilon}_{i})
,\quad v_i\in\mathcal{V}_{a},\ v_j\in\mathcal{V}_{i}^{r},
\end{equation}
where $\mathcal{V}_{i}^{r}$ denotes the randomly sampled interaction targets of node $v_i$, and $\eta(\delta)$ denotes a difficulty-dependent stochastic perturbation factor. As $\delta$ increases, the activation probability of feature perturbation increases, replacing the obvious zero-feature pattern with attribute variations that are harder to distinguish from normal patterns.

\textbf{Community Dispersion (CD).}
This type simulates anomalies such as scalper accounts, whose interactions are scattered across different user or product groups without a clear community preference. Specifically, we connect each anomalous node to a randomly sampled set of nodes across the graph and perturb its attributes around the global feature mean:
\begin{equation}
a_{ij}^{s}=1,\qquad
\mathbf{x}_{i}^{s}
=\tilde{\mathbf{x}}
+
\lambda(\delta)\boldsymbol{\epsilon}_{i},
\quad
v_i\in\mathcal{V}_{a},\ v_j\in\mathcal{V}_{i}^{r},
\end{equation}
where $\tilde{\mathbf{x}}$ denotes the global feature mean, $\mathcal{V}_{i}^{r}$ is the randomly sampled target set, and $\lambda(\delta)$ controls the feature deviation. Increasing $\delta$ decreases $\lambda(\delta)$, bringing anomalous features closer to the global feature pattern, while increasing the probability of additional feature perturbation and cross-community rewiring to introduce more complex structural and attribute variations.

\textbf{Mixed anomaly.}
Real-world graphs may contain multiple anomaly patterns simultaneously rather than a single type. Therefore, under the mixed-anomaly setting, we partition the anomalous node set $\mathcal{V}_{a}$ into several subsets and apply different injection strategies to them, allowing multiple anomaly types to coexist within the same graph.

By organizing the three synthesis stages into a unified ``forge'', \oursyn flexibly configures and combines graph structures, features, and anomaly patterns, covering a broad range of anomalous graph scenarios. Their flexible combinations can produce a virtually unlimited variety of anomalous graphs, providing rich training data for generalist GAD models.

\section{Details of Topological Pattern Encoding}
\label{app:topo}

In Eq.~\ref{eq:topology}, $f_{\mathrm{top}}(\cdot)$ summarizes five anomaly-aware topological statistics into the topological descriptor $\mathbf{r}_i$. Here, we provide their detailed construction.

\noindent\textbf{Global cosine similarity} measures the directional consistency between a node feature and the global mean feature. We first obtain the graph-propagated representation
$\hat{\mathbf{X}}=\hat{\mathbf{A}}^{2}\mathbf{X}$, where
$\hat{\mathbf{A}}=\tilde{\mathbf{D}}^{-\frac{1}{2}}
\tilde{\mathbf{A}}\tilde{\mathbf{D}}^{-\frac{1}{2}}$
and $\tilde{\mathbf{A}}=\mathbf{A}+\mathbf{I}$.
Then, the global cosine similarity is computed as
\begin{equation}
r_i^{\mathrm{g}}
=
\frac{
\hat{\mathbf{x}}_i^{\top}\mathbf{c}_{g}
}{
\|\hat{\mathbf{x}}_i\|_2\|\mathbf{c}_{g}\|_2
},
\qquad
\mathbf{c}_{g}
=
\frac{1}{N}\sum_{j=1}^{N}\hat{\mathbf{x}}_j .
\end{equation}

\noindent\textbf{Neighborhood cosine similarity} measures the directional consistency between a node and its neighbors:
\begin{equation}
r_i^{\mathrm{nc}}
=
\frac{1}{|\mathcal{N}_i|}
\sum_{v_j\in\mathcal{N}_i}
\frac{
\hat{\mathbf{x}}_i^{\top}\hat{\mathbf{x}}_j
}{
\|\hat{\mathbf{x}}_i\|_2
\|\hat{\mathbf{x}}_j\|_2
},
\end{equation}
where $\mathcal{N}_i$ denotes the one-hop neighbors of node $v_i$.

\noindent\textbf{Neighborhood Euclidean distance} measures the positional discrepancy between a node and its neighbors in the representation space:
\begin{equation}
r_i^{\mathrm{ne}}
=
\frac{1}{|\mathcal{N}_i|}
\sum_{v_j\in\mathcal{N}_i}
\left\|
\hat{\mathbf{x}}_i-\hat{\mathbf{x}}_j
\right\|_2.
\end{equation}

\noindent\textbf{\textit{Degree}} characterizes the connectivity of node $v_i$:
\begin{equation}
d_i =
\sum_{j=1}^{N}a_{ij}.
\end{equation}

\noindent\textbf{Local clustering coefficient} measures the connectivity density among the neighbors of node $v_i$:
\begin{equation}
r_i^{\mathrm{l}}
=
\frac{2T_i}
{d_i(d_i-1)},
\end{equation}
where $T_i$ denotes the number of triangles containing $v_i$, and $r_i^{\mathrm{l}}=0$ when $d_i<2$.

Together, these five statistics characterize node patterns from global consistency, neighborhood consistency, and structural connectivity. To reduce cross-graph scale variations, we independently apply rank normalization to each statistic. Specifically, the rank-based transformation is defined as 
$
\mathcal{R}(m_i)
=
\frac{\operatorname{rank}(m_i)}{N},
$
where $m \in \{r^{\mathrm{g}}, r^{\mathrm{nc}}, r^{\mathrm{ne}}, d, r^{\mathrm{l}} \}$ denotes the relational attribute, and $\text{rank}(m_i)$ is the ascending rank index of node $v_i$ within the dataset.
The topological descriptor in Eq.~\ref{eq:topology} is therefore constructed as
\begin{equation}
\mathbf{r}_i
=
\left[
\mathcal{R}(r_i^{\mathrm{g}}),
\mathcal{R}(r_i^{\mathrm{nc}}),
\mathcal{R}(r_i^{\mathrm{ne}}),
\mathcal{R}(d_i),
\mathcal{R}(r_i^{\mathrm{l}})
\right].
\end{equation}

\section{Complexity Analysis}
We analyze the computational complexity of \ourmethod in two stages: topology-semantic feature construction and anomaly inference.
In the feature construction stage, the topological patterns are obtained through node-wise structural statistics followed by rank transformation, with an approximate complexity of $\mathcal{O}(|\mathcal{E}|D+n\log n)$ on sparse graphs, where $n$, $|\mathcal{E}|$, and $D$ denote the numbers of nodes, edges, and input feature dimensions, respectively. For semantic encoding, PCA, landmark-based scattering, and MLP encoding result in a complexity of approximately \(\mathcal{O}(nDd' + nd'^2)\), since the number of landmarks is much smaller than the number of nodes.
During anomaly inference, the two streams project node representations into the hidden space and compare each query node with a support set of size $k$. The resulting complexity is $\mathcal{O}(nd'^2 + nkd')$. Since $k$ and $d'$ are typically much smaller than $n$, the inference cost grows approximately linearly with the number of nodes. Table~\ref{tab:inference_time} further provides an empirical comparison of the average inference time over the 14 real-world datasets. \ourmethod achieves an average inference time of 1.96 seconds, which is comparable to ReFi-GAD and substantially faster than IA-GGAD and UNPrompt.

\begin{table}[htbp]
    \centering
    \caption{Average inference time across 14 real-world datasets.}
    \label{tab:inference_time}
    \begin{tabular}{lccccc}
        \toprule
        \textbf{Method} & ARC & UNPrompt & IA-GGAD & ReFi-GAD & Ours  \\
        \midrule
        \textbf{Avg. Time (s)} & 1.4 & 5.62 & 3.34 & 1.88 & 1.96  \\
        \bottomrule
    \end{tabular}
\end{table}

\section{Experimental Details}

\subsection{Datasets}\label{appe:data}
We evaluate our method on 14 real-world graph datasets covering four application domains: citation, social, e-commerce, and behavioral/interaction networks. These datasets vary substantially in graph scale, feature dimensionality, connectivity patterns, and anomaly ratios, providing diverse environments for evaluating cross-domain generalization. Detailed statistics are reported in Table~\ref{tab:dataset_statistics}, followed by descriptions of each dataset:

\begin{itemize}[leftmargin=10pt, itemsep=0pt, parsep=0pt, topsep=0pt]
    \item[$\bullet$] \textbf{Citeseer}~\cite{sen2008collective} is a citation network where nodes represent scientific publications and edges denote citation relationships. Node features are bag-of-words representations of paper content. 
 
    \item[$\bullet$] \textbf{CS} (Coauthor-CS)~\cite{shchur2018pitfalls} is an academic collaboration network where nodes represent authors and edges indicate co-authorship relations. Node features are derived from keywords of the authors' publications. 
 
    \item[$\bullet$] \textbf{ACM}~\cite{tang2008arnetminer} is a citation network where nodes represent research papers and edges denote citation relationships. Node features are derived from paper abstracts. 
 
    \item[$\bullet$] \textbf{BlogCatalog}~\cite{ding2019deep} is a social network where nodes represent bloggers and edges denote social relationships. Node features are tags describing users and their blogs. 
 
    \item[$\bullet$] \textbf{Amazon}~\cite{mcauley2013amateurs} is a user co-review network where nodes represent users and edges connect users reviewing the same products. Node features characterize users' review behaviors and statistics. 
 
    \item[$\bullet$] \textbf{Photo} (Photo)~\cite{shchur2018pitfalls} is a product co-purchase network where nodes represent products and edges connect frequently co-purchased products. Node features are extracted from product reviews. 
 
    \item[$\bullet$] \textbf{Weibo}~\cite{kumar2019predicting} is a social network where nodes represent users and edges connect users sharing hashtags. Node features contain user location information and posted textual content. 
 
    \item[$\bullet$] \textbf{Cora}~\cite{sen2008collective} is a citation network where nodes represent scientific publications and edges denote citation relationships. Node features are bag-of-words representations of paper content. 
 
    \item[$\bullet$] \textbf{Pubmed}~\cite{sen2008collective} is a biomedical citation network where nodes represent scientific articles and edges correspond to citation relationships. Node features are TF--IDF representations of paper content. 
 
    \item[$\bullet$] \textbf{Flickr}~\cite{tang2009relational} is a social network where nodes represent users and edges denote social relationships. Node features are tags associated with users' shared photos. 
 
    \item[$\bullet$] \textbf{Facebook}~\cite{xu2022contrastive} is a social network where nodes represent users and edges denote social connections. Node features consist of user attributes. 
 
    \item[$\bullet$] \textbf{YelpChi}~\cite{rayana2015collective} is a review network where nodes represent reviews and edges connect reviews posted by the same user. Node features characterize review behaviors and statistics. 
 
    \item[$\bullet$] \textbf{Questions}~\cite{platonov2023critical} is a user interaction network where nodes represent users and edges indicate question--answer interactions. Node features are derived from users' textual profile descriptions. 
 
    \item[$\bullet$] \textbf{Reddit}~\cite{kumar2019predicting} is a social interaction network where nodes represent users and edges connect users interacting under the same posts. Node features are derived from users' posted textual content.
\end{itemize}

\begin{table*}[htbp]
  \centering
  \caption{The statistics of datasets organized by application domains.}
  \label{tab:dataset_statistics}
    \begin{tabular}{l|cccccc}
    \toprule
    Dataset & \#Nodes & \#Edges & \#Features & Avg. Degree & \#Anomaly & \%Anomaly \\
    \midrule
\rowcolor{gray!10}

    \multicolumn{7}{c}{Citation Networks} \\
    Cora & 2,708 & 11,604 & 1,433 & 4.29 & 150 & 5.54 \\
    Citeseer & 3,327 & 10,154 & 3,703 & 3.05 & 150 & 4.51 \\
    ACM & 16,484 & 147,866 & 8,337 & 8.97 & 597 & 3.62 \\
    Pubmed & 19,717 & 92,846 & 500 & 4.71 & 600 & 3.04 \\
    CS & 18,333 & 167,986 & 6,805 & 9.16 & 600 & 3.27 \\
    \midrule
\rowcolor{gray!10}

    \multicolumn{7}{c}{Social Networks} \\
    BlogCatalog & 5,196 & 345,558 & 8,189 & 66.50 & 300 & 5.77 \\
    Flickr & 7,575 & 482,602 & 12,047 & 63.71 & 450 & 5.94 \\
    Facebook & 1,081 & 55,104 & 576 & 50.98 & 25 & 2.31 \\
    \midrule
\rowcolor{gray!10} 

    \multicolumn{7}{c}{E-commerce Networks} \\
    Amazon & 10,224 & 351,216 & 25 & 34.35 & 693 & 6.78 \\
    YelpChi & 23,831 & 98,630 & 32 & 4.14 & 1,217 & 5.11 \\
    Photo & 7,650 & 241,306 & 745 & 31.54 & 450 & 5.88 \\
    \midrule
    \rowcolor{gray!10} 
    \multicolumn{7}{c}{Behavioral/Interaction Networks} \\
    Weibo & 8,405 & 407,963 & 400 & 48.54 & 868 & 10.33 \\
    Questions & 48,921 & 153,540 & 301 & 3.14 & 1,460 & 2.98 \\
    Reddit & 10,984 & 157,032 & 64 & 14.30 & 366 & 3.33 \\
    \bottomrule
    \end{tabular}%
\end{table*}

\subsection{Baselines}\label{appe:baseline}

\noindent\textbf{GNN Baselines}
\begin{itemize}[leftmargin=10pt, itemsep=0pt, parsep=0pt, topsep=0pt]
    \item[$\bullet$] \textbf{GCN}~\cite{kipf2016semi} is a classical graph neural network that learns node representations by aggregating neighborhood information through graph convolutions. We employ it as a basic supervised GNN baseline for node anomaly classification.

    \item[$\bullet$] \textbf{GAT}~\cite{velivckovic2017graph} introduces an attention mechanism to assign different importance weights to neighboring nodes during message aggregation.
\end{itemize}

\noindent\textbf{Unsupervised GAD Methods}
\begin{itemize}[leftmargin=10pt, itemsep=0pt, parsep=0pt, topsep=0pt]
    \item[$\bullet$] \textbf{CoLA}~\cite{liu2021anomaly} employs contrastive self-supervised learning to measure the agreement between a target node and its local subgraph, and derives anomaly scores from their contrastive consistency.

    \item[$\bullet$] \textbf{ANEMONE}~\cite{jin2021anemone} extends contrastive GAD to multiple graph scales by jointly modeling node--subgraph agreement at the patch and context levels.

    \item[$\bullet$] \textbf{SmoothGNN}~\cite{dong2025smoothgnn} exploits the different smoothing behaviors of normal and anomalous nodes and models both individual and neighborhood smoothing patterns for unsupervised anomaly detection.
\end{itemize}

\noindent\textbf{Supervised GAD Methods}
\begin{itemize}[leftmargin=10pt, itemsep=0pt, parsep=0pt, topsep=0pt]
    \item[$\bullet$] \textbf{AHFAN}~\cite{wang2025graph} learns hybrid node representations through frequency-aware graph filtering and attention-based feature modeling to improve anomaly discrimination.

    \item[$\bullet$] \textbf{BWGNN}~\cite{tang2022rethinking} analyzes graph anomalies from the spectral perspective and employs beta-wavelet band-pass filters to capture the high-frequency signals associated with anomalous nodes.

    \item[$\bullet$] \textbf{BGNN}~\cite{ivanov2021boost} jointly combines gradient-boosted decision trees with graph neural networks, enabling the model to capture both heterogeneous node attributes and graph structural information.

    \item[$\bullet$] \textbf{GHRN}~\cite{gao2023addressing} addresses graph heterophily from a spectral perspective by identifying and suppressing potentially inter-class edges, reducing the interference of abnormal neighborhood connections.

    \item[$\bullet$] \textbf{GGAD}~\cite{qiao2024generative} studies semi-supervised GAD by generating pseudo anomalies from known normal nodes and using them as informative negative samples for discriminative anomaly detection.
\end{itemize}

\noindent\textbf{Generalist GAD Methods}
\begin{itemize}[leftmargin=10pt, itemsep=0pt, parsep=0pt, topsep=0pt]
    \item[$\bullet$] \textbf{ARC}~\cite{liu2024arc} introduces in-context learning for generalist GAD, aligning heterogeneous graph features and leveraging a small set of target-domain normal samples to detect anomalies on unseen graphs without retraining.

    \item[$\bullet$] \textbf{UNPrompt}~\cite{niu2024zero} develops unified neighborhood prompts and uses latent node attribute predictability as a generalized anomaly measure, enabling zero-shot anomaly detection across heterogeneous graphs.

    \item[$\bullet$] \textbf{IA-GGAD}~\cite{zhang2026ia} addresses feature-space and graph-structure shifts through invariant representation learning and structure-insensitive affinity learning for zero-shot cross-domain anomaly detection.

    \item[$\bullet$] \textbf{ReFi-GAD}~\cite{liu2026rethinking} constructs relational fingerprints to align heterogeneous graph features and combines a Transformer-based encoder with SNR-guided refinement for transferable anomaly detection across unseen graphs.
\end{itemize}

\subsection{Experimental Details}\label{appe:implement}

\noindent\textbf{Metrics}
We employ two widely used metrics for graph anomaly detection: Area Under the Receiver Operating Characteristic Curve (\textbf{AUROC}) and Area Under the Precision-Recall Curve (\textbf{AUPRC}). AUROC measures the ranking quality between anomalous and normal nodes, while AUPRC evaluates the trade-off between precision and recall. Higher values indicate better performance for both metrics. We repeat each experiment five times and report the mean and standard deviation.

\noindent\textbf{Implementation Pipeline}
Our method is trained exclusively on synthetic graphs generated by \oursyn and directly evaluated on real-world target graphs. For existing generalist GAD methods, following previous work~\cite{liu2026rethinking}, we divide the 14 real-world datasets into two disjoint and domain-balanced groups. Models are trained on one group and evaluated on the other, after which the two groups are swapped to ensure that every target graph remains unseen during training.

To further evaluate the quality of synthetic data, we additionally train these generalist GAD methods on graphs generated by \oursyn and compare them with their counterparts trained on real-world source graphs. For \ourmethod, the topological encoder uses a hidden embedding dimension of $d'=256$ and 4 Transformer layers, and the batch size is 512. The default support size is $K=30$ and is kept consistent across all few-shot generalist methods. All methods use consistent hyperparameters across target datasets, without parameter updates or dataset-specific tuning during testing.

\section{Additional Experimental Results}
\subsection{Comparison Experiments}\label{appe:main_exp}
We provide the complete comparison results of all methods in this subsection. Table~\ref{tab:main_auroc} presents the detailed AUROC results with standard deviations, and Table~\ref{tab:fullmain_auprc} further reports the AUPRC results with standard deviations.

First, \oursyn$+$\ourmethod achieves the best AUROC performance on all 14 real-world datasets, demonstrating strong and stable generalization across different graph domains. Compared with existing generalist GAD methods, \ourmethod shows a clear performance advantage, indicating that the large-scale synthetic graphs generated by \oursyn provide sufficient transferable anomaly knowledge, while the proposed topology-semantic coordinated architecture and curriculum learning further enable the model to effectively exploit such knowledge. Meanwhile, existing generalist GAD methods trained on \oursyn-generated graphs achieve performance and rankings broadly comparable to their counterparts trained on real-world graphs, further demonstrating the effectiveness of the proposed synthetic data generation strategy.

Second, \ourmethod achieves the best AUPRC performance on 10 out of 14 datasets and ranks within the top three on 11 datasets, showing a clear overall advantage. In particular, \ourmethod achieves the best results across citation, social, e-commerce, and behavioral graphs, demonstrating that its performance advantage is not limited to a specific graph domain. Overall, the \oursyn-trained generalist GAD models also remain competitive with their real-world-trained counterparts, further supporting the effectiveness of \oursyn in providing useful training data for generalist GAD.

\input{tables/appendix_auroc}
\input{tables/appendix_auprc}

\subsection{Ablation Study of \oursyn.}\label{appe:ablation_data}
Table~\ref{tab:ablation_data} presents the complete ablation results of different generation strategies in \oursyn.
For structure generation, removing CS, SF, PB, and SW leads to average AUROC drops of 3.54\%, 2.34\%, 2.03\%, and 1.39\%, respectively. The performance degradation is particularly pronounced when CS is removed, indicating that community-structured graphs provide important topological patterns for transferable anomaly learning. Moreover, among the three generation stages, structure ablation causes the largest average performance drop, further highlighting the importance of topological diversity.
For feature generation, removing Dense, Sparse, and Binary features results in average performance drops of 1.04\%, 2.05\%, and 2.32\%, respectively. This shows that different feature types provide complementary semantic environments and improve the model's adaptability to heterogeneous feature patterns.
For anomaly injection, removing LD, LC, AS, and CD decreases the average AUROC by 0.95\%, 1.07\%, 0.96\%, and 1.28\%, respectively. Although the impact varies across datasets, all four anomaly mechanisms contribute to better overall performance. These results indicate that incorporating anomaly patterns with diverse structural and attribute characteristics and varying detection difficulties helps the model learn more transferable anomaly knowledge and improves its performance across diverse real-world graphs.

\subsection{Ablation Study of \ourmethod.}\label{appe:ablation_model}
Table~\ref{tab:ablation_model} presents the complete ablation results of different components in \ourmethod. The results show that all components contribute to the overall performance and complement each other.
First, removing the topological stream (\textbf{- Topo}) reduces the average AUROC from 86.61\% to 66.13\%, a substantial drop of 20.48\%. This highlights the importance of topological encoding in capturing structural evidence of anomalous behaviors.
Second, removing the semantic stream (\textbf{- Attr}) decreases the average AUROC by 4.58\%, with particularly large drops on Amazon and Photo. This indicates that semantic encoding provides complementary anomaly evidence when topological patterns alone are insufficient.
Moreover, removing curriculum learning (\textbf{- Curr}) leads to an average AUROC drop of 5.43\%, with degradation on all 14 datasets. This shows that progressive learning helps the model to first establish stable anomaly decision boundaries and then improve their reliability with increasingly difficult anomaly patterns.
Finally, replacing the target-aware gating mechanism with simple average fusion (\textbf{- Gat}) decreases the average AUROC by 0.97\%. This demonstrates that Anomaly Evidence Coordination effectively coordinates topological and semantic evidence according to node-specific topological patterns.

\input{tables/app_ablation_syn}
\input{tables/app_ablation_model}

\subsection{Additional Parameter Sensitivity Analysis}
\label{app:sensitivity}

\begin{figure}[b]
    \centering
    \vspace{-2mm}
    \includegraphics[
        width=0.35\textwidth,
        trim=0 0 0 20,
        clip
    ]{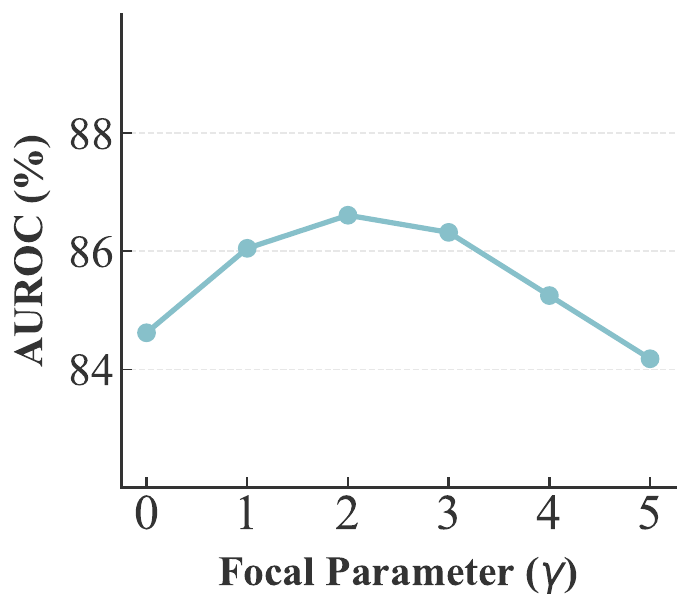}
    \vspace{-2mm}
    \caption{Sensitivity to $\gamma$ of \ourmethod.}
    \label{fig.parameter_gamma}
    \vspace{-3mm}
\end{figure}

We investigate the sensitivity of \ourmethod to the focal parameter $\gamma$ by varying $\gamma$ within $\{0, 1, \dots, 5\}$. The experimental results are illustrated in Fig.~\ref{fig.parameter_gamma}. As shown in the figure, the performance first improves as $\gamma$ increases and reaches the best AUROC of 86.61\% at $\gamma=2$, after which it gradually declines. When $\gamma$ is too small, the focal mechanism provides insufficient emphasis on difficult samples that have not been well fitted, limiting the model's ability to learn challenging anomaly patterns. When $\gamma$ is too large, the model may overemphasize these difficult samples and consequently suffer from overfitting, thereby reducing its generalization ability.

\subsection{Additional Visualizations}
\label{app:visualization}

\begin{figure}[t]
    \centering
    \includegraphics[
        width=0.8\textwidth,
        trim=0 0 0 0,
        clip
    ]{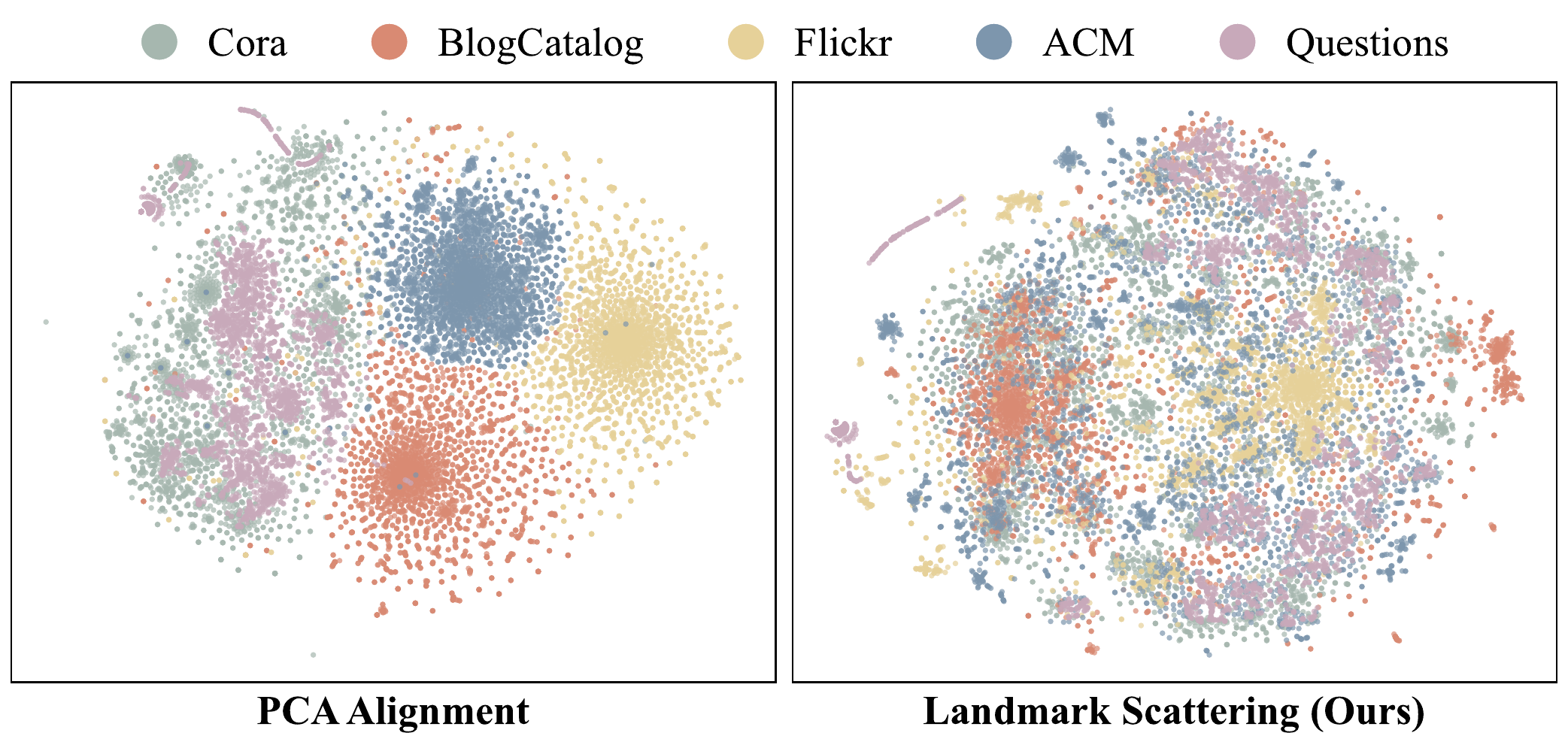}
    \vspace{-2mm}
    \caption{Visualization of node embeddings.}
    \label{fig.alignment}
\end{figure}

\textbf{Visualization of landmark-based scattering.} 
To evaluate the effectiveness of landmark scattering for cross-domain feature alignment, we compare the feature representations produced by PCA and landmark scattering on several real-world graph datasets. For visualization, we apply simple centering and orientation alignment to the representations from different datasets. 
As shown in Fig.~\ref{fig.alignment}, after PCA, the representations from different datasets are still clearly separated, showing that dimensional unification alone cannot effectively reduce cross-domain feature differences. In contrast, landmark scattering leads to much larger overlap among different datasets while still preserving local structures. This shows that our alignment strategy can further reduce cross-domain feature discrepancies and provide a more consistent semantic space for the shared projector and scoring model.

\begin{figure}[t]
    \centering
    \begin{minipage}{0.9\linewidth}
        \centering

        \begin{subfigure}{0.48\linewidth}
            \centering
            \includegraphics[width=\linewidth]{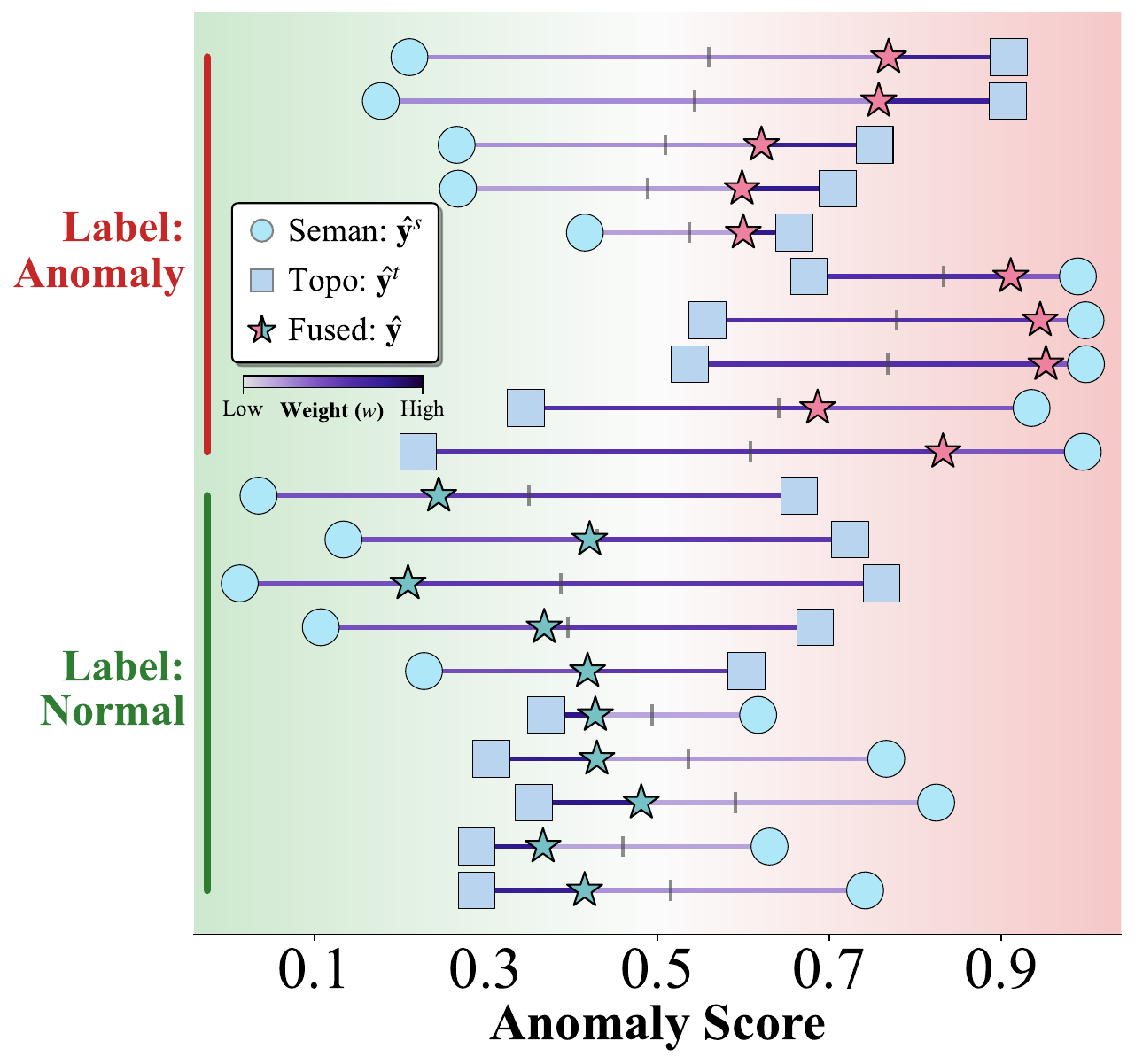}
            \caption{Synthetic training}
            \label{fig:Gating1}
        \end{subfigure}
        \hfill
        \begin{subfigure}{0.48\linewidth}
            \centering
            \includegraphics[width=\linewidth]{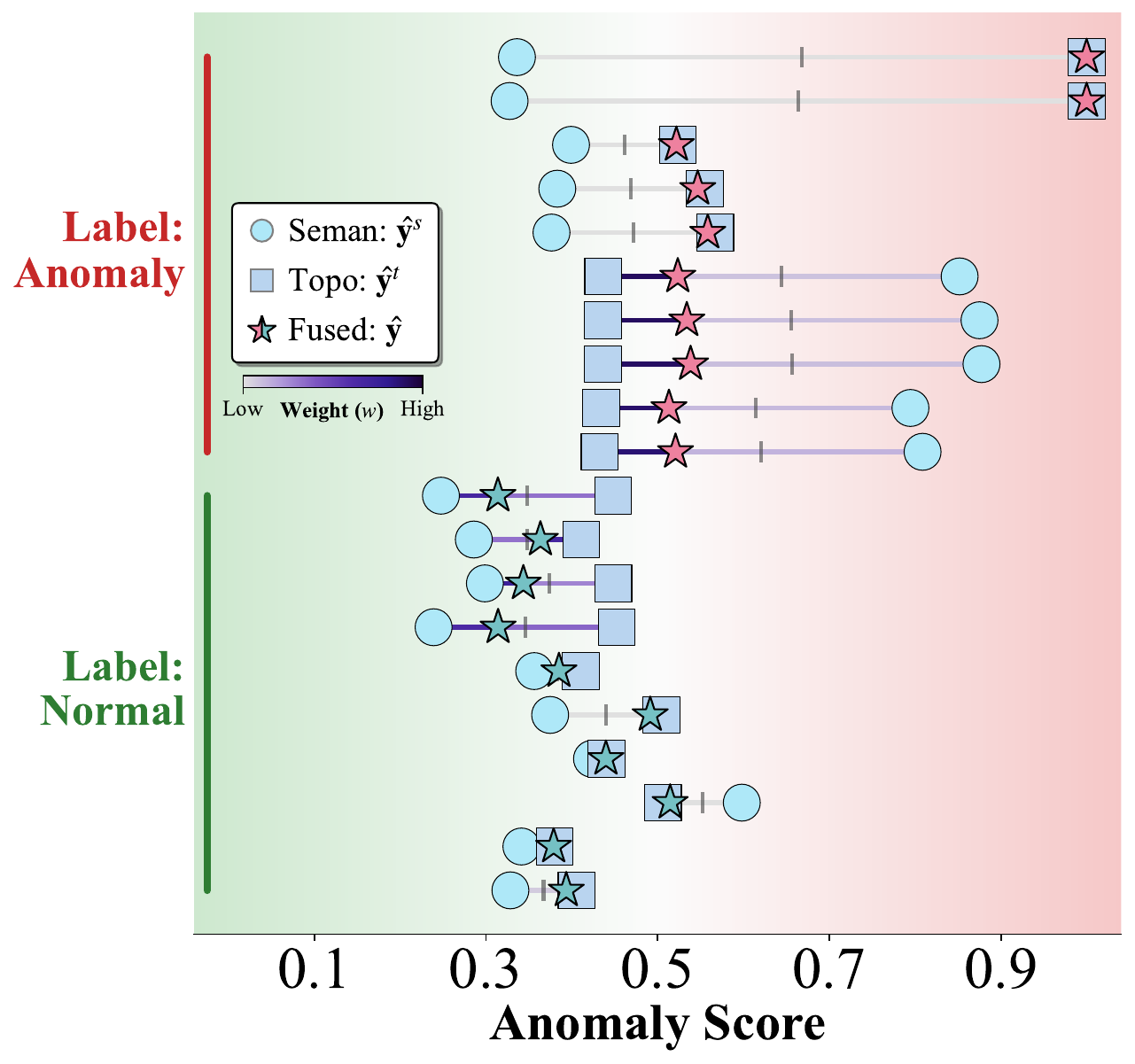}
            \caption{Real-world training}
            \label{fig:Gating2}
        \end{subfigure}

        \vspace{-2mm}
        \caption{Visualization of anomaly evidence coordination.}
        \label{fig:Gating}
    \end{minipage}
    \vspace{-3mm}
\end{figure}
\noindent\textbf{Visualization of Anomaly Evidence Coordination.} To further examine the behavior of our gating strategy, we sample several nodes from Amazon for visualization. As shown in Fig.~\ref{fig:Gating1}, our gating module adaptively coordinates the contributions of the topological and semantic streams for different samples, leading to more accurate anomaly predictions. We further conduct the same analysis using a model trained only on real-world data. As shown in Fig.~\ref{fig:Gating2}, this model tends to assign higher weights to the topological stream for many samples and shows less flexible coordination between the two streams. One possible reason is that the limited real-world training data provide insufficiently diverse anomaly patterns, making the gating module more likely to rely on the dominant topological signals observed during training. In contrast, large-scale and diverse synthetic data provide richer supervision for learning the gating strategy, enabling more effective coordination of the two streams. This further shows the value of synthetic data in alleviating the training data bottleneck of generalist GAD.

\subsection{Examples of Generated Synthetic Graphs}
\label{app:syn_examples}
To provide a more concrete view of the synthetic data generated by \oursyn, Table~\ref{tab:syn_examples} reports the configurations and statistics of the first 20 generated graphs. These graphs represent only a subset of the 108 synthetic graphs used for model training. They cover different topology types, feature types, anomaly mechanisms, and difficulty levels, while also varying in graph size, feature dimensionality, and anomaly ratio, illustrating the diversity of the synthetic training corpus produced by \oursyn.

\begin{table*}[t]
\centering
\caption{Configurations and statistics of example synthetic graphs generated by \oursyn.}
\label{tab:syn_examples}
\small
\setlength{\tabcolsep}{5pt}
\resizebox{\textwidth}{!}{
\begin{tabular}{c c c c c r r r r}
\toprule
ID & Structure & Feature & Anomaly & Difficulty
& $N$ & $D$ & $N_a$ & Ratio (\%) \\
\midrule
0  & CS & Sparse & LC  & 0.3 & 13,058 & 1,292 & 956 & 7.32 \\
1  & SF & Sparse & LC  & 0.7 & 8,151  & 1,048 & 619 & 7.59 \\
2  & CS & Dense  & Mix & 0.9 & 12,888 & 560   & 813 & 6.31 \\
3  & PB & Binary & Mix & 0.9 & 12,549 & 217   & 891 & 7.10 \\
4  & SW & Binary & AS  & 0.1 & 11,443 & 1,888 & 649 & 5.67 \\
5  & SW & Dense  & Mix & 0.5 & 10,720 & 1,619 & 755 & 7.04 \\
6  & SF & Binary & LD  & 0.1 & 7,785  & 1,099 & 562 & 7.22 \\
7  & CS & Binary & LC  & 0.1 & 14,283 & 1,268 & 892 & 6.25 \\
8  & PB & Sparse & LC  & 0.1 & 11,354 & 687   & 892 & 7.86 \\
9  & SW & Binary & Mix & 0.5 & 7,480  & 1,746 & 397 & 5.31 \\
10 & CS & Dense  & LC  & 0.7 & 12,094 & 1,914 & 612 & 5.06 \\
11 & PB & Sparse & LD  & 0.9 & 3,739  & 300   & 261 & 6.98 \\
12 & SF & Sparse & AS  & 0.1 & 9,965  & 539   & 776 & 7.79 \\
13 & SW & Binary & CD  & 0.9 & 13,647 & 1,736 & 749 & 5.49 \\
14 & CS & Sparse & LC  & 0.7 & 3,953  & 1,670 & 223 & 5.64 \\
15 & SW & Dense  & AS  & 0.5 & 14,105 & 1,400 & 868 & 6.15 \\
16 & SF & Sparse & AS  & 0.1 & 9,001  & 1,155 & 554 & 6.15 \\
17 & CS & Sparse & CD  & 0.5 & 11,632 & 1,697 & 873 & 7.51 \\
18 & SW & Binary & Mix & 0.3 & 13,616 & 828   & 982 & 7.21 \\
19 & CS & Binary & CD  & 0.3 & 9,657  & 869   & 703 & 7.28 \\
\bottomrule
\end{tabular}}
\end{table*}

%% file: tables/appendix_auroc.tex
\begin{table*}[!t]
  \centering
  \caption{Anomaly detection performance in terms of AUROC (\%) $\pm$ std. Highlighted are the results ranked \first{first}, \second{second}, and \third{third}. \colorbox{agforgePurple}{Purple-shaded rows} denote models trained with \oursyn-generated data.}
  \resizebox{\textwidth}{!}{
    \begin{tabular}{l|ccccccc|c}
    \toprule
    \multirow{1}[1]{*}{Method}
          & Citeseer & CS    & ACM   & BlogCatalog & Amazon & Photo & Weibo &  Rank \\
          \midrule
    \rowcolor[rgb]{ .906,  .902,  .902} \multicolumn{9}{c}{GAD Methods} \\
    GCN   & 48.32 $\pm$ 1.03 & 56.19 $\pm$ 1.23 & 50.43 $\pm$ 1.04 & 43.06 $\pm$ 0.92 & 58.69 $\pm$ 0.17 & 47.70 $\pm$ 2.10 & 46.40 $\pm$ 1.82 & 16.29 \\
    GAT   & 63.11 $\pm$ 1.92 & 59.25 $\pm$ 0.94 & 61.48 $\pm$ 0.94 & 60.50 $\pm$ 0.73 & 54.84 $\pm$ 2.68 & 45.63 $\pm$ 1.39 & 73.51 $\pm$ 2.66 & 14.00 \\
    CoLA  & 73.81 $\pm$ 2.87 & 65.99 $\pm$ 2.30 & 54.94 $\pm$ 4.14 & 60.58 $\pm$ 5.07 & 65.02 $\pm$ 10.86 & 65.18 $\pm$ 1.52 & 41.08 $\pm$ 7.06 & 12.57 \\
    SmoothGNN & 90.42 $\pm$ 8.81 & 78.65 $\pm$ 5.25 & 78.00 $\pm$ 4.92 & 73.06 $\pm$ 1.91 & 50.75 $\pm$ 3.01 & 44.80 $\pm$ 5.02 & 85.96 $\pm$ 4.50 & 10.00 \\
    ANEMONE & 41.36 $\pm$ 2.40 & 42.18 $\pm$ 1.19 & 45.53 $\pm$ 3.24 & 35.18 $\pm$ 3.43 & 43.51 $\pm$ 4.29 & 49.22 $\pm$ 4.87 & 35.57 $\pm$ 4.75 & 18.29 \\
    AHFAN & 48.02 $\pm$ 1.74 & 59.09 $\pm$ 0.63 & 63.70 $\pm$ 1.21 & 59.60 $\pm$ 0.25 & 29.50 $\pm$ 2.60 & 52.16 $\pm$ 0.60 & 63.03 $\pm$ 5.92 & 15.29 \\
    BWGNN & 60.91 $\pm$ 1.50 & 59.12 $\pm$ 1.92 & 61.80 $\pm$ 1.54 & 67.26 $\pm$ 1.55 & 50.40 $\pm$ 5.49 & 60.82 $\pm$ 2.63 & 81.62 $\pm$ 2.44 & 13.00 \\
    BGNN  & 55.14 $\pm$ 3.74 & 52.14 $\pm$ 0.86 & 52.01 $\pm$ 1.08 & 52.65 $\pm$ 1.09 & 52.17 $\pm$ 2.43 & 47.26 $\pm$ 0.87 & 77.47 $\pm$ 4.03 & 15.57 \\
    GHRN  & 74.25 $\pm$ 2.39 & 78.01 $\pm$ 0.75 & 76.88 $\pm$ 0.50 & 65.09 $\pm$ 0.51 & 52.93 $\pm$ 1.06 & 51.99 $\pm$ 1.19 & 57.77 $\pm$ 2.23 & 11.57 \\
    GGAD  & 67.03 $\pm$ 4.99 & 64.76 $\pm$ 2.20 & 67.02 $\pm$ 0.64 & 58.83 $\pm$ 0.07 & 50.36 $\pm$ 4.58 & 58.72 $\pm$ 1.78 & 70.23 $\pm$ 0.77 & 13.57 \\
    \midrule
    \rowcolor[rgb]{ .906,  .902,  .902} \multicolumn{9}{c}{Generalist GAD Methods} \\
    ARC   & 89.96 $\pm$ 0.27 & 82.73 $\pm$ 0.25 & 79.83 $\pm$ 0.11 & 73.98 $\pm$ 0.44 & \second{79.77} $\pm$ 1.69 & 75.54 $\pm$ 0.34 & 88.05 $\pm$ 0.40 & 5.43 \\
    \rowcolor[rgb]{0.93, 0.90, 0.97} +\oursyn   & 89.70 $\pm$ 0.25 & 82.99 $\pm$ 0.36 & 79.94 $\pm$ 0.16 & 74.16 $\pm$ 0.38 & 71.23 $\pm$ 2.18 & \third{75.67} $\pm$ 0.53 & 87.70 $\pm$ 0.29 & 5.43 \\
    \midrule
    UNPrompt & 72.59 $\pm$ 2.50 & 74.38 $\pm$ 1.19 & 73.99 $\pm$ 0.94 & 69.02 $\pm$ 0.24 & 63.14 $\pm$ 11.27 & 69.61 $\pm$ 2.04 & 49.19 $\pm$ 1.34 & 11.14 \\
    \rowcolor[rgb]{0.93, 0.90, 0.97} +\oursyn   & 76.82 $\pm$ 2.27 & 76.11 $\pm$ 1.37 & 74.55 $\pm$ 0.89 & 69.18 $\pm$ 0.59 & 62.98 $\pm$ 9.79 & 71.06 $\pm$ 2.43 & 49.37 $\pm$ 0.72 & 10.00 \\
    \midrule
    IA-GGAD & 91.84 $\pm$ 0.16 & 93.04 $\pm$ 0.33 & 91.19 $\pm$ 0.33 & \third{74.76} $\pm$ 0.57 & \third{71.93} $\pm$ 8.43 & 70.23 $\pm$ 1.47 & \third{91.10} $\pm$ 0.22 & 4.29 \\
    \rowcolor[rgb]{0.93, 0.90, 0.97} +\oursyn   & 92.18 $\pm$ 0.31 & 91.49 $\pm$ 0.47 & 90.14 $\pm$ 0.31 & \second{74.81} $\pm$ 0.52 & 70.87 $\pm$ 4.11 & 72.50 $\pm$ 1.16 & \second{91.55} $\pm$ 0.14 & 4.14 \\
    \midrule
    ReFi-GAD & \third{95.60} $\pm$ 0.94 & \third{98.69} $\pm$ 0.20 & \third{93.94} $\pm$ 0.69 & 69.96 $\pm$ 2.62 & 65.98 $\pm$ 1.14 & 73.78 $\pm$ 1.37 & 88.52 $\pm$ 3.10 & 4.71 \\
    \rowcolor[rgb]{0.93, 0.90, 0.97} +\oursyn   & \second{96.40} $\pm$ 1.85 & \second{99.03} $\pm$ 0.20 & \second{93.96} $\pm$ 1.70 & 69.42 $\pm$ 3.89 & 68.32 $\pm$ 1.83 & \second{76.58} $\pm$ 2.81 & 89.02 $\pm$ 2.80 & 3.71 \\
    \midrule
    \rowcolor[rgb]{0.93, 0.90, 0.97} \ourmethod & \first{96.52} $\pm$ 0.57 & \first{99.43} $\pm$ 0.11 & \first{96.24} $\pm$ 0.44 & \first{76.19} $\pm$ 1.65 & \first{87.10} $\pm$ 2.35 & \first{82.15} $\pm$ 2.27 & \first{93.53} $\pm$ 0.62 & \first{1.00} \\
    \midrule
    \multicolumn{1}{c}{} &       &       &       &       &       &       & \multicolumn{1}{c}{} &  \\
    \midrule
    \multirow{1}[1]{*}{Method} 
          & Cora  & Pubmed & Flickr & Facebook & YelpChi & Questions & Reddit & Rank \\
          \midrule
    \rowcolor[rgb]{ .906,  .902,  .902} \multicolumn{9}{c}{GAD Methods} \\
    GCN   & 32.41 $\pm$ 1.23 & 33.68 $\pm$ 0.65 & 38.07 $\pm$ 0.78 & 76.35 $\pm$ 0.56 & 51.21 $\pm$ 2.23 & 41.37 $\pm$ 0.67 & 46.80 $\pm$ 0.75 & 15.86 \\
    GAT   & 56.07 $\pm$ 1.08 & 67.21 $\pm$ 0.75 & 56.05 $\pm$ 1.30 & 55.33 $\pm$ 1.91 & 50.43 $\pm$ 2.08 & 57.37 $\pm$ 0.45 & 43.39 $\pm$ 0.41 & 15.29 \\
    CoLA  & 66.02 $\pm$ 2.06 & 70.09 $\pm$ 2.94 & 60.64 $\pm$ 2.84 & 70.88 $\pm$ 5.50 & 52.41 $\pm$ 1.55 & 53.23 $\pm$ 4.66 & 50.60 $\pm$ 1.27 & 12.29 \\
    SmoothGNN & 88.39 $\pm$ 5.98 & 78.22 $\pm$ 6.62 & 76.95 $\pm$ 4.07 & 53.62 $\pm$ 8.98 & 61.30 $\pm$ 6.52 & 59.01 $\pm$ 4.83 & 54.09 $\pm$ 2.43 & 7.29 \\
    ANEMONE & 47.08 $\pm$ 0.37 & 37.65 $\pm$ 0.61 & 37.07 $\pm$ 1.14 & 45.55 $\pm$ 3.60 & 50.88 $\pm$ 2.53 & 51.55 $\pm$ 1.16 & 52.41 $\pm$ 2.42 & 16.29 \\
    AHFAN & 55.89 $\pm$ 1.16 & 64.08 $\pm$ 7.14 & 62.77 $\pm$ 8.24 & 32.11 $\pm$ 6.72 & 49.76 $\pm$ 2.34 & 54.01 $\pm$ 3.36 & 46.90 $\pm$ 0.36 & 16.00 \\
    BWGNN & 63.62 $\pm$ 1.82 & 68.19 $\pm$ 0.62 & 68.37 $\pm$ 2.71 & 41.55 $\pm$ 2.56 & 52.76 $\pm$ 1.62 & 57.59 $\pm$ 3.49 & \second{60.24} $\pm$ 1.12 & 10.29 \\
    BGNN  & 52.66 $\pm$ 4.09 & 58.46 $\pm$ 2.59 & 52.56 $\pm$ 1.57 & 65.80 $\pm$ 14.26 & 50.67 $\pm$ 5.75 & 56.80 $\pm$ 2.62 & 41.80 $\pm$ 2.13 & 15.57 \\
    GHRN  & 68.18 $\pm$ 2.51 & 75.15 $\pm$ 1.30 & 64.32 $\pm$ 0.90 & 33.07 $\pm$ 2.16 & 51.72 $\pm$ 0.58 & \second{59.81} $\pm$ 1.44 & 53.45 $\pm$ 0.72 & 11.00 \\
    GGAD  & 71.15 $\pm$ 0.42 & 69.82 $\pm$ 0.20 & 59.79 $\pm$ 0.84 & 20.24 $\pm$ 0.73 & 52.14 $\pm$ 3.75 & 58.92 $\pm$ 0.86 & 53.78 $\pm$ 1.25 & 12.00 \\
    \midrule
    \rowcolor[rgb]{ .906,  .902,  .902} \multicolumn{9}{c}{Generalist GAD Methods} \\
    ARC   & 83.59 $\pm$ 0.70 & 84.44 $\pm$ 0.27 & 75.61 $\pm$ 0.97 & 61.11 $\pm$ 2.31 & 54.47 $\pm$ 1.33 & \third{59.48} $\pm$ 0.47 & 58.94 $\pm$ 0.36 & 6.29 \\
    \rowcolor[rgb]{0.93, 0.90, 0.97} +\oursyn   & 85.19 $\pm$ 0.35 & 85.21 $\pm$ 0.13 & 74.48 $\pm$ 0.31 & 72.09 $\pm$ 1.02 & 50.56 $\pm$ 0.66 & 58.70 $\pm$ 0.34 & \third{59.40} $\pm$ 0.65 & 7.71 \\
    \midrule
    UNPrompt & 71.23 $\pm$ 2.76 & 82.48 $\pm$ 0.55 & 69.68 $\pm$ 0.29 & 76.34 $\pm$ 5.54 & 52.36 $\pm$ 3.75 & 46.17 $\pm$ 1.22 & 56.14 $\pm$ 1.90 & 9.57 \\
    \rowcolor[rgb]{0.93, 0.90, 0.97} +\oursyn   & 67.47 $\pm$ 3.08 & 81.78 $\pm$ 1.23 & 69.64 $\pm$ 0.16 & 74.48 $\pm$ 4.65 & 50.63 $\pm$ 3.21 & 45.27 $\pm$ 1.23 & 56.24 $\pm$ 1.70 & 11.43 \\
    \midrule
    IA-GGAD & 87.90 $\pm$ 0.47 & 87.84 $\pm$ 0.45 & 69.05 $\pm$ 0.95 & 79.86 $\pm$ 2.78 & 52.38 $\pm$ 1.22 & 57.82 $\pm$ 0.34 & 57.97 $\pm$ 1.28 & 6.57 \\
    \rowcolor[rgb]{0.93, 0.90, 0.97} +\oursyn   & 87.83 $\pm$ 0.34 & 88.56 $\pm$ 0.53 & 67.04 $\pm$ 1.02 & 80.70 $\pm$ 2.15 & 51.38 $\pm$ 1.07 & 58.22 $\pm$ 0.21 & 57.84 $\pm$ 0.92 & 7.29 \\
    \midrule
    ReFi-GAD & \second{95.74} $\pm$ 1.54 & \second{97.07} $\pm$ 0.50 & \second{81.25} $\pm$ 4.88 & \second{92.79} $\pm$ 2.06 & \third{68.90} $\pm$ 2.29 & 57.47 $\pm$ 3.07 & 57.40 $\pm$ 2.93 & 4.29 \\
    \rowcolor[rgb]{0.93, 0.90, 0.97} +\oursyn   & \third{94.56} $\pm$ 1.43 & \third{96.66} $\pm$ 0.60 & \third{77.28} $\pm$ 2.15 & \third{90.80} $\pm$ 3.30 & \second{73.65} $\pm$ 0.92 & 57.78 $\pm$ 2.62 & 58.74 $\pm$ 3.59 & 4.00 \\
    \midrule
    \rowcolor[rgb]{0.93, 0.90, 0.97} \ourmethod & \first{96.41} $\pm$ 1.48 & \first{97.77} $\pm$ 0.55 & \first{86.46} $\pm$ 1.45 & \first{97.41} $\pm$ 1.17 & \first{74.46} $\pm$ 0.72 & \first{64.62} $\pm$ 2.32 & \first{64.31} $\pm$ 2.75 & \first{1.00} \\
    \bottomrule
    \end{tabular}}%
  \label{tab:main_auroc}%
\end{table*}

%% file: tables/appendix_auprc.tex
\begin{table*}[!t]
  \centering
  \caption{Anomaly detection performance in terms of AUPRC (\%) $\pm$ std. Highlighted are the results ranked \first{first}, \second{second}, and \third{third}. \colorbox{agforgePurple}{Purple-shaded rows} denote models trained with \oursyn-generated data.}
  \resizebox{\textwidth}{!}{
    \begin{tabular}{l|ccccccc|c}
    \toprule
    \multirow{1}[1]{*}{Method} 
          & Citeseer & CS    & ACM   & BlogCatalog & Amazon & Photo & Weibo & Rank \\
          \midrule
    \rowcolor[rgb]{ .906,  .902,  .902} \multicolumn{9}{c}{GAD Methods} \\
    GCN   & 7.63 $\pm$ 2.98 & 8.35 $\pm$ 3.53 & 7.10 $\pm$ 3.33 & 4.95 $\pm$ 0.16 & 8.52 $\pm$ 0.16 & 5.48 $\pm$ 0.39 & 9.95 $\pm$ 1.01 & 15.71 \\
    GAT   & 7.37 $\pm$ 0.61 & 6.85 $\pm$ 0.53 & 8.06 $\pm$ 0.55 & 13.29 $\pm$ 0.55 & 7.95 $\pm$ 0.45 & 4.92 $\pm$ 0.14 & 53.35 $\pm$ 3.83 & 14.29 \\
    CoLA  & 15.34 $\pm$ 2.30 & 21.09 $\pm$ 3.70 & 18.16 $\pm$ 2.39 & 25.23 $\pm$ 2.36 & 12.99 $\pm$ 4.17 & 10.98 $\pm$ 1.56 & 24.57 $\pm$ 6.31 & 9.71 \\
    SmoothGNN & 41.39 $\pm$ 13.91 & 17.06 $\pm$ 3.50 & 16.81 $\pm$ 4.82 & 20.33 $\pm$ 2.75 & 8.70 $\pm$ 2.24 & 5.16 $\pm$ 0.73 & 43.60 $\pm$ 7.87 & 12.14 \\
    ANEMONE & 4.43 $\pm$ 0.77 & 3.18 $\pm$ 0.25 & 3.58 $\pm$ 0.17 & 4.54 $\pm$ 0.28 & 6.15 $\pm$ 0.52 & 6.76 $\pm$ 0.53 & 9.25 $\pm$ 0.84 & 18.29 \\
    AHFAN & 45.16 $\pm$ 2.64 & 57.14 $\pm$ 1.59 & \second{61.29} $\pm$ 1.05 & \first{56.92} $\pm$ 0.36 & 28.43 $\pm$ 5.27 & \first{50.29} $\pm$ 1.25 & 57.48 $\pm$ 7.19 & 4.00 \\
    BWGNN & 5.85 $\pm$ 0.43 & 4.09 $\pm$ 0.35 & 5.26 $\pm$ 0.44 & 10.24 $\pm$ 0.75 & 6.27 $\pm$ 0.67 & 7.97 $\pm$ 0.56 & 44.19 $\pm$ 6.11 & 15.43 \\
    BGNN  & 5.12 $\pm$ 0.63 & 4.45 $\pm$ 0.56 & 4.45 $\pm$ 0.40 & 6.42 $\pm$ 0.18 & 7.35 $\pm$ 0.46 & 5.16 $\pm$ 0.11 & 62.69 $\pm$ 6.41 & 15.43 \\
    GHRN  & 15.01 $\pm$ 2.10 & 15.17 $\pm$ 3.57 & 23.57 $\pm$ 0.77 & 14.76 $\pm$ 0.84 & 7.38 $\pm$ 0.38 & 6.02 $\pm$ 0.29 & 11.80 $\pm$ 1.00 & 13.00 \\
    GGAD  & 11.12 $\pm$ 3.14 & 5.64 $\pm$ 0.43 & 7.34 $\pm$ 0.46 & 7.67 $\pm$ 0.02 & 7.10 $\pm$ 1.69 & 7.47 $\pm$ 0.45 & 16.23 $\pm$ 0.20 & 15.14 \\
    \midrule
    \rowcolor[rgb]{ .906,  .902,  .902} \multicolumn{9}{c}{Generalist GAD Methods} \\
    ARC   & 47.56 $\pm$ 0.58 & 36.39 $\pm$ 0.14 & 40.79 $\pm$ 0.13 & 35.17 $\pm$ 0.22 & \third{44.42} $\pm$ 1.17 & \third{30.25} $\pm$ 1.11 & 51.03 $\pm$ 8.69 & 4.86 \\
    \rowcolor[rgb]{0.93, 0.90, 0.97} +\oursyn   & 45.36 $\pm$ 0.37 & 36.11 $\pm$ 0.17 & 39.70 $\pm$ 0.05 & \second{36.30} $\pm$ 0.13 & 15.22 $\pm$ 2.88 & \second{32.10} $\pm$ 0.90 & 60.72 $\pm$ 0.38 & 5.43 \\
    \midrule
    UNPrompt & 10.41 $\pm$ 1.85 & 18.30 $\pm$ 3.06 & 16.95 $\pm$ 0.64 & 28.90 $\pm$ 1.79 & 12.73 $\pm$ 5.85 & 16.34 $\pm$ 5.96 & 18.78 $\pm$ 2.16 & 10.14 \\
    \rowcolor[rgb]{0.93, 0.90, 0.97} +\oursyn   & 13.41 $\pm$ 2.22 & 18.42 $\pm$ 3.02 & 16.87 $\pm$ 0.97 & 28.67 $\pm$ 1.80 & 11.48 $\pm$ 4.48 & 18.42 $\pm$ 6.18 & 18.19 $\pm$ 2.16 & 10.29 \\
    \midrule
    IA-GGAD & 48.82 $\pm$ 0.68 & 37.68 $\pm$ 1.21 & \third{47.82} $\pm$ 0.45 & 34.98 $\pm$ 0.10 & \second{50.79} $\pm$ 2.90 & 14.61 $\pm$ 1.27 & \third{71.61} $\pm$ 0.28 & 4.57 \\
    \rowcolor[rgb]{0.93, 0.90, 0.97} +\oursyn   & 50.34 $\pm$ 0.70 & 37.25 $\pm$ 0.46 & 45.77 $\pm$ 0.39 & \third{36.09} $\pm$ 0.25 & 12.23 $\pm$ 3.13 & 13.39 $\pm$ 1.27 & \second{71.80} $\pm$ 0.41 & 5.71 \\
    \midrule
    ReFi-GAD & \second{67.05} $\pm$ 7.71 & \third{59.31} $\pm$ 5.73 & 46.44 $\pm$ 5.95 & 13.09 $\pm$ 1.08 & 13.84 $\pm$ 1.13 & 16.68 $\pm$ 4.14 & 61.77 $\pm$ 6.79 & 5.71 \\
    \rowcolor[rgb]{0.93, 0.90, 0.97} +\oursyn   & \third{58.01} $\pm$ 3.37 & \second{82.85} $\pm$ 4.03 & 39.00 $\pm$ 9.76 & 11.12 $\pm$ 1.88 & 12.28 $\pm$ 0.81 & 14.60 $\pm$ 5.50 & 55.24 $\pm$ 8.99 & 7.57 \\
    \midrule
    \rowcolor[rgb]{0.93, 0.90, 0.97} \ourmethod  & \first{76.44} $\pm$ 2.13 & \first{87.69} $\pm$ 2.47 & \first{62.75} $\pm$ 9.67 & 22.97 $\pm$ 2.03 & \first{53.96} $\pm$ 11.71 & 27.09 $\pm$ 5.05 & \first{75.04} $\pm$ 3.22 & \first{2.57} \\
    \midrule
    \multicolumn{1}{c}{} &       &       &       &       &       &       & \multicolumn{1}{c}{} &  \\
    \midrule
    \multirow{1}[1]{*}{Method} 
          & Cora  & Pubmed & Flickr & Facebook & YelpChi & Questions & Reddit & Rank \\
          \midrule
    \rowcolor[rgb]{ .906,  .902,  .902} \multicolumn{9}{c}{GAD Methods} \\
    GCN   & 3.95 $\pm$ 0.14 & 2.18 $\pm$ 0.06 & 4.52 $\pm$ 0.03 & 4.95 $\pm$ 0.20 & 6.33 $\pm$ 0.37 & 2.47 $\pm$ 0.05 & 3.16 $\pm$ 0.03 & 16.00 \\
    GAT   & 6.28 $\pm$ 0.35 & 4.44 $\pm$ 0.12 & 12.19 $\pm$ 0.78 & 2.46 $\pm$ 0.12 & 6.14 $\pm$ 0.57 & 3.50 $\pm$ 0.11 & 2.67 $\pm$ 0.02 & 14.86 \\
    CoLA  & 10.97 $\pm$ 1.13 & 17.41 $\pm$ 2.09 & 23.03 $\pm$ 3.50 & 7.35 $\pm$ 2.02 & 5.67 $\pm$ 0.14 & 3.45 $\pm$ 0.58 & 3.53 $\pm$ 0.22 & 11.71 \\
    SmoothGNN & 43.34 $\pm$ 9.10 & 13.28 $\pm$ 3.21 & 20.18 $\pm$ 0.74 & 4.57 $\pm$ 4.33 & 7.94 $\pm$ 2.17 & 4.69 $\pm$ 0.78 & 3.62 $\pm$ 0.32 & 9.29 \\
    ANEMONE & 5.47 $\pm$ 0.42 & 2.48 $\pm$ 0.06 & 4.84 $\pm$ 0.13 & 2.57 $\pm$ 0.52 & 5.24 $\pm$ 0.46 & 3.16 $\pm$ 0.06 & 3.63 $\pm$ 0.26 & 16.57 \\
    AHFAN & 6.25 $\pm$ 0.12 & 5.78 $\pm$ 0.10 & 15.12 $\pm$ 0.32 & 9.09 $\pm$ 0.24 & 6.10 $\pm$ 0.17 & \first{7.31} $\pm$ 0.16 & 2.95 $\pm$ 0.12 & 11.00 \\
    BWGNN & 11.16 $\pm$ 2.33 & 7.33 $\pm$ 1.06 & 15.40 $\pm$ 1.57 & 2.19 $\pm$ 0.43 & 5.48 $\pm$ 0.19 & 4.30 $\pm$ 0.42 & \second{5.43} $\pm$ 0.44 & 11.57 \\
    BGNN  & 5.76 $\pm$ 0.96 & 3.58 $\pm$ 0.28 & 8.15 $\pm$ 0.49 & 7.09 $\pm$ 3.75 & 6.27 $\pm$ 0.95 & 3.84 $\pm$ 0.69 & 2.66 $\pm$ 0.15 & 14.00 \\
    GHRN  & 15.28 $\pm$ 1.70 & 10.38 $\pm$ 0.62 & 17.50 $\pm$ 1.64 & 1.70 $\pm$ 0.07 & 5.00 $\pm$ 0.13 & 4.50 $\pm$ 0.46 & 4.66 $\pm$ 0.48 & 11.43 \\
    GGAD  & 15.12 $\pm$ 0.38 & 5.95 $\pm$ 0.04 & 7.42 $\pm$ 0.16 & 1.41 $\pm$ 0.01 & 6.86 $\pm$ 0.87 & 4.32 $\pm$ 0.06 & 4.57 $\pm$ 0.13 & 11.43 \\
    \midrule
    \rowcolor[rgb]{ .906,  .902,  .902} \multicolumn{9}{c}{Generalist GAD Methods} \\
    ARC   & 47.04 $\pm$ 0.37 & 32.14 $\pm$ 0.52 & \first{39.59} $\pm$ 0.12 & 7.04 $\pm$ 1.69 & 6.23 $\pm$ 0.08 & 4.16 $\pm$ 0.09 & 4.93 $\pm$ 0.15 & 6.71 \\
    \rowcolor[rgb]{0.93, 0.90, 0.97} +\oursyn   & 46.33 $\pm$ 1.14 & 32.46 $\pm$ 0.48 & \second{39.51} $\pm$ 0.18 & 5.99 $\pm$ 1.15 & 5.83 $\pm$ 0.18 & 4.23 $\pm$ 0.08 & 4.63 $\pm$ 0.10 & 7.57 \\
    \midrule
    UNPrompt & 11.15 $\pm$ 1.08 & 14.14 $\pm$ 0.94 & 23.67 $\pm$ 1.86 & 8.09 $\pm$ 2.96 & 6.62 $\pm$ 1.13 & 2.97 $\pm$ 0.17 & 3.82 $\pm$ 0.16 & 10.14 \\
    \rowcolor[rgb]{0.93, 0.90, 0.97} +\oursyn   & 9.40 $\pm$ 0.94 & 13.00 $\pm$ 2.51 & 24.64 $\pm$ 1.26 & 8.41 $\pm$ 2.73 & 5.65 $\pm$ 0.74 & 3.13 $\pm$ 0.35 & 4.06 $\pm$ 0.40 & 11.57 \\
    \midrule
    IA-GGAD & 46.89 $\pm$ 1.07 & 32.27 $\pm$ 2.22 & \third{37.26} $\pm$ 0.14 & 5.59 $\pm$ 0.50 & 5.98 $\pm$ 0.19 & 4.38 $\pm$ 0.11 & 4.56 $\pm$ 0.19 & 7.71 \\
    \rowcolor[rgb]{0.93, 0.90, 0.97} +\oursyn   & 47.46 $\pm$ 0.79 & 35.23 $\pm$ 1.83 & 36.96 $\pm$ 0.58 & 5.93 $\pm$ 0.52 & 5.79 $\pm$ 0.30 & 4.21 $\pm$ 0.08 & 4.34 $\pm$ 0.18 & 8.29 \\
    \midrule
    ReFi-GAD & \second{64.03} $\pm$ 7.62 & \third{49.50} $\pm$ 8.55 & 27.57 $\pm$ 5.04 & \third{20.51} $\pm$ 6.51 & \third{8.83} $\pm$ 1.42 & 4.81 $\pm$ 0.71 & 4.47 $\pm$ 0.43 & 4.29 \\
    \rowcolor[rgb]{0.93, 0.90, 0.97} +\oursyn   & \third{50.14} $\pm$ 7.57 & \second{60.03} $\pm$ 8.75 & 14.46 $\pm$ 2.69 & \second{24.53} $\pm$ 13.95 & \second{11.57} $\pm$ 1.18 & \third{5.21} $\pm$ 0.52 & \third{4.96} $\pm$ 0.78 & 4.14 \\
    \midrule
    \rowcolor[rgb]{0.93, 0.90, 0.97} \ourmethod  & \first{79.25} $\pm$ 2.92 & \first{70.50} $\pm$ 10.09 & 33.87 $\pm$ 5.73 & \first{56.10} $\pm$ 13.42 & \first{11.86} $\pm$ 0.94 & \second{6.74} $\pm$ 0.81 & \first{6.22} $\pm$ 1.52 & \first{1.71} \\
    \bottomrule
    \end{tabular}}%
  \label{tab:fullmain_auprc}%
\end{table*}

%% file: tables/app_ablation_syn.tex
\begin{table*}[!t]
  \centering
  \caption{Ablation study of different components in \oursyn. Performance is evaluated in terms of AUROC (\%).}
  \normalsize
  
  \setlength{\tabcolsep}{3pt} 
  \newcolumntype{Y}{>{\centering\arraybackslash}p{0.9cm}}

  \resizebox{\textwidth}{!}{
    \begin{tabular}{l|*{14}{Y}|c}
    \toprule
    {Variant} 
          & Cite & CS    & ACM   & Blog & Amz & Photo & Weibo & Cora  & Pubmed & Flickr & FB & Yelp & Quest & Reddit & {Average} \\
    \midrule
    \rowcolor[rgb]{ .906,  .902,  .902} \multicolumn{16}{c}{Structure Generation} \\
    w/o CS & 96.17 & 99.11 & 95.03 & 73.64 & 71.36 & 74.47 & 90.79 & 95.77 & 97.19 & 80.04 & 96.64 & 74.60 & 60.47 & 57.75 & 83.07 \\
    w/o SF & 95.27 & 99.20 & 94.69 & 71.71 & 83.78 & 76.77 & 90.16 & 95.88 & 97.21 & 76.81 & 96.90 & 74.44 & 64.78 & 62.24 & 84.27 \\
    w/o PB & 96.37 & 98.71 & 95.08 & 72.72 & 81.40 & 76.81 & 92.43 & 96.14 & 97.54 & 79.43 & 96.13 & 74.71 & 64.06 & 62.52 & 84.58 \\
    w/o SW & 95.23 & 99.17 & 95.63 & 74.53 & 84.28 & 77.91 & 92.32 & 95.91 & 97.34 & 81.83 & 97.15 & 74.75 & 63.86 & 63.13 & 85.22 \\
    \midrule
    \rowcolor[rgb]{ .906,  .902,  .902} \multicolumn{16}{c}{Feature Generation} \\
    w/o Dense & 96.51 & 98.93 & 94.04 & 75.73 & 83.44 & 80.69 & 93.81 & 95.51 & 97.23 & 82.07 & 98.02 & 76.31 & 63.09 & 62.66 & 85.57 \\
    w/o Sparse & 96.82 & 98.87 & 96.19 & 72.79 & 75.32 & 77.11 & 91.58 & 96.61 & 97.89 & 83.38 & 97.58 & 75.87 & 63.15 & 60.64 & 84.56 \\
    w/o Binary & 95.87 & 99.36 & 95.28 & 74.53 & 79.07 & 73.89 & 93.07 & 95.78 & 97.53 & 80.75 & 96.26 & 75.25 & 63.70 & 59.76 & 84.29 \\
    \midrule
    \rowcolor[rgb]{ .906,  .902,  .902} \multicolumn{16}{c}{Anomaly Injection} \\
    w/o LD & 95.15 & 98.56 & 94.76 & 72.22 & 84.99 & 81.55 & 93.30 & 94.84 & 96.73 & 84.63 & 97.23 & 76.91 & 64.97 & 63.38 & 85.66 \\
    w/o LC & 95.49 & 99.44 & 94.56 & 73.69 & 89.58 & 80.32 & 92.43 & 94.75 & 97.44 & 79.98 & 96.08 & 74.17 & 66.02 & 63.58 & 85.54 \\
    w/o AS & 96.73 & 99.08 & 95.53 & 75.98 & 88.59 & 78.07 & 92.46 & 96.87 & 97.78 & 80.01 & 96.05 & 75.01 & 64.78 & 62.17 & 85.65 \\
    w/o CD & 97.20 & 99.16 & 95.03 & 75.21 & 86.46 & 79.24 & 93.59 & 96.11 & 96.35 & 80.45 & 97.66 & 72.59 & 63.46 & 62.09 & 85.33 \\
    \midrule
    \rowcolor[rgb]{0.93, 0.90, 0.97} \oursyn (Ours) & 96.52 & 99.43 & 96.24 & 76.19 & 87.10 & 82.15 & 93.53 & 96.41 & 97.77 & 86.46 & 97.41 & 74.46 & 64.62 & 64.31 & 86.61 \\
    \bottomrule
    \end{tabular}}%
  \label{tab:ablation_data}%
\end{table*}

%% file: tables/app_ablation_model.tex
\begin{table*}[!t]
  \centering
  \caption{Ablation study of different components in \ourmethod. Performance is evaluated in terms of AUROC (\%).}
  \normalsize
  
  \setlength{\tabcolsep}{3pt} 
  \newcolumntype{Y}{>{\centering\arraybackslash}p{0.9cm}}

  \resizebox{\textwidth}{!}{
    \begin{tabular}{l|*{14}{Y}|c}
    \toprule
    {Variant} 
          & Cite & CS    & ACM   & Blog & Amz & Photo & Weibo & Cora  & Pubmed & Flickr & FB & Yelp & Quest & Reddit & {Average} \\
    \midrule
    - Topo & 62.34 & 63.25 & 57.01 & 64.37 & 91.04 & 69.88 & 83.50 & 64.16 & 57.59 & 58.80 & 69.12 & 57.91 & 62.19 & 64.63 & 66.13 \\
    - Attr & 95.43 & 98.95 & 95.24 & 71.90 & 68.06 & 70.75 & 89.25 & 95.64 & 97.06 & 79.09 & 95.51 & 74.61 & 59.79 & 57.18 & 82.03 \\
    - Curr & 95.60 & 98.69 & 93.94 & 69.96 & 65.98 & 73.78 & 88.52 & 95.06 & 97.28 & 81.55 & 92.79 & 69.20 & 56.91 & 57.21 & 81.18 \\
    - Gat  & 95.84 & 98.92 & 94.58 & 75.47 & 90.25 & 79.25 & 92.43 & 95.50 & 96.96 & 79.14 & 96.20 & 73.90 & 65.55 & 64.90 & 85.64 \\
    \midrule
    \rowcolor[rgb]{0.93, 0.90, 0.97} Ours & 96.52 & 99.43 & 96.24 & 76.19 & 87.10 & 82.15 & 93.53 & 96.41 & 97.77 & 86.46 & 97.41 & 74.46 & 64.62 & 64.31 & 86.61 \\
    \bottomrule
    \end{tabular}}%
  \label{tab:ablation_model}%
\end{table*}